# An Integrated Inverse Space Sparse Representation Framework for Tumor Classification

Xiaohui Yang[1*], Wenming Wu[1], Yunmei Chen[2], Xianqi Li[2], Juan Zhang[3], Dan Long[3], Lijun Yang[1]

1. Data Analysis Technology Lab, Institute of Applied Mathematics, Henan University, Kaifeng 475004, China
2. Department of Mathematics, University of Florida, Gainesville, Florida, USA
3. Zhejiang Cancer Hospital, Hangzhou 310022, China

**Abstract:** Microarray gene expression data-based tumor classification is an active and challenging issue. In this paper, an integrated tumor classification framework is presented, which aims to exploit information in existing available samples, and focuses on the small sample problem and unbalanced classification problem. Firstly, an inverse space sparse representation based classification (ISSRC) model is proposed by considering the characteristics of gene-based tumor data, such as sparsity and a small number of training samples. A decision information factors (DIF)-based gene selection method is constructed to enhance the representation ability of the ISSRC. It is worth noting that the DIF is established from reducing clinical misdiagnosis rate and dimension of small sample data. For further improving the representation ability and classification stability of the ISSRC, feature learning is conducted on the selected gene subset. The feature learning method is constructed by complementing the advantages of non-negative matrix factorization (NMF) and deep learning. Without confusion, the ISSRC combined with gene selection and feature learning is called the integrated ISSRC, whose stability, optimization and the corresponding convergence are analyzed. Extensive experiments on six public microarray gene expression datasets show the integrated ISSRC-based tumor classification framework is superior to classical and state-of-the-art methods. There are significant improvements in classification accuracy, specificity and sensitivity, whether there is a tumor in the early diagnosis, what kind of tumor, or whether metastasis occurs after tumor surgery.

***Key words:*** Tumor classification, microarray gene expression data, decision information genes, layer-wise pre-training sparse NMF, inverse space sparse representation.

# 1. Introduction

Microarray technology with its ability to simultaneously interrogate 10,000–40,000 genes has changed people's thinking of molecular classification of tumors [1]. In general, tumor recognition involves three levels: early diagnosis, tumor type recognition, and whether cancer metastasis occurs after surgery. It's necessary to effectively explore and analyze tumor pathogenesis from the molecular biology aspects [2, 3].

___________

*Corresponding author
Email addresses: xhyanghenu@163.com (X. H. Yang), wenmingwu55@163.com (W. M. Wu), yun@ufl.edu (Y. M. Chen), xli1124@gmail.com (X. Q. Li), zhangjuan_496@163.com (J. Zhang), legend_long@aliyun.com (D. Long), yanglijun@henu.edu.cn (L.J. Yang).

Effective tumor classification plays an important role in clinical diagnosis and treatment. Classifier design is a critical issue for tumor classification. Commonly used classification methods for microarray gene expression data are random forest [4], neural networks [5], support vector machine (SVM) [6], etc. Most of these methods have been developed on statistical learning theory, which relies on model parameters and may produce "over-fitting". Sparse representation is a sparse coding technique based on an over-completed dictionary. Sparse representation based classification (SRC) was originally proposed and used for face recognition [7]. SRC achieves good result when there are sufficient training samples per subject. Recently, SRC and its improved methods have been widely used in microarray gene expression data-based tumor classification [8-12]. Zheng et al. [12] made use of singular value decomposition to learn a dictionary and then classified gene expression data of tumor subtypes based on SRC. However, it is difficult to acquire sufficient and effective labeled samples for tumor classification. In addition, Zhang et al. [13] indicated that the discrimination ability of SRC will be reduced when there is a small disturbance on the representation error. In our previous work [14], an inverse projection-based pseudo-full-space representation classification (PFSRC) method was proposed for face recognition. PFSRC made full use of complementary information between existing face samples. Microarray gene expression data, however, have no obvious complementary information similar to faces. Training samples from other categories may lead to interference information rather than complementary information. Therefore, it is important to utilize the characteristics of the gene data. Unfortunately, microarray gene expression data have the characteristics of small samples (patients), high dimensions (thousands of genes) and high redundancy [15], which impose a great challenge to tumor classification.

As a dimension reduction method for small sample problem, gene selection aims to remove irrelevant, redundant genes and obtain a small set of information genes [16]. Therefore, an effective gene selection method may enhance the representation ability for small sample problem. The general gene selection methods can be classified into three categories: filters [17, 18], embedded [19] and wrappers [20, 21] methods. As a filter method, Dudoit [17] proposed a between-groups to within-groups sum of squares (BW) method, which is simple and stable. Algamal et al. [19] proposed a sparse logistic regression-based embedded method. Embedded gene selection methods, however, combine gene selection and classification in an optimal process, where classifier training may weaken the ability of gene selection to a certain extent. Moreover, implementation and computation process of the embedded methods are always complex. Ruiz et al. [20] proposed a wrapper method based on statistical significance. Xie et al. [21] proposed a differentially expressed gene selection algorithm for unbalanced gene datasets by maximize the area under curve (AUC), where the curve is receive operating characteristic curve (ROC). ROC exhibits the accuracy of a binary classifier as its discrimination threshold varies [22]. The larger the AUC is, the better the classifier is. However, ROC only focuses on AUC without taking into account that misdiagnosis rate and missed diagnosis rate, while the clinic is more concerned with the latter. Decision curve analysis (DCA) [23] is just a way of evaluating treatment plans by maximizing the clinic net benefit (NB) of profit minuses harm. DCA evaluates a treatment plan by risk (of illness) – (clinic) NB ratio, which aims to select treatments corresponding to low clinical misdiagnosis rate. It is undoubted that there is more practical value by integrating clinical needs into gene selection-based tumor classification.

Feature learning can further explore the more essential information contained in the selected information gene subset. NMF [24] is a feature learning method that does not rely on category information, and can explore useful information contained in all available samples simultaneously, even if there are only a small number of training samples. In recent years, NMF and its improved methods have achieved good results in many fields [25-31]. Hoyer [32] proposed a sparse NMF (SNMF). Zheng et al. [29] used SNMF to perform gene selection and tumor classification by combining SVM. However, NMF methods are affected by the initial value of the iteration. Deep learning is a popular feature representation learning method [33]. Some preliminary results in recognizing benign and malignant tumor have been obtained [34, 35]. However, the success of deep learning relies on the available large-scale training data, complex network structures, high-performance GPU devices and optimized parallel algorithms. As a data-driven feature learning method deep learning relies on large number of effective training samples. Tumor classification, however, is a typical small sample problem. Xu and Sun et al. proposed a model-driven deep learning method [36] to complement the advantages between the model and the data. It is interesting and promising that different feature representation learning approaches complement each other.

From the viewpoint of optimization, SRC methods belong to an underdetermined linear system. To alleviate this problem, different constraints have been introduced by considering priori information. In the field of microarray gene expression data-based tumor recognition, sparsity is important prior information. The sparsity embodies $l_0$-regularization constraint and can be relaxed to $l_1$-regularization. Commonly used methods for solving $l_1$-regularization problems are least angle regression (LARS) [37] and the alternating direction method of multipliers (ADMM) [38, 39]. For biostatistics, ADMM has attracted a great deal of attention because it mainly deals with convex optimization problems with constraints. Xiao et al. [39] proposed a generalized ADMM with semi-proximal terms, denoted as GsADMM, which is competitive to the classic ADMM in terms of the convergence error and the convergence speed.

Motivated by these works, a tumor classification framework is proposed based on an integrated inverse space sparse representation classification (ISSRC) model, whose performance is further enhanced by integrating gene selection and feature learning. It is noted that the integrated ISSR model focuses on utilizing the existing available samples to alleviate small sample problem and classification stability problem. The main contributions are as follows and shown in Fig. 1.

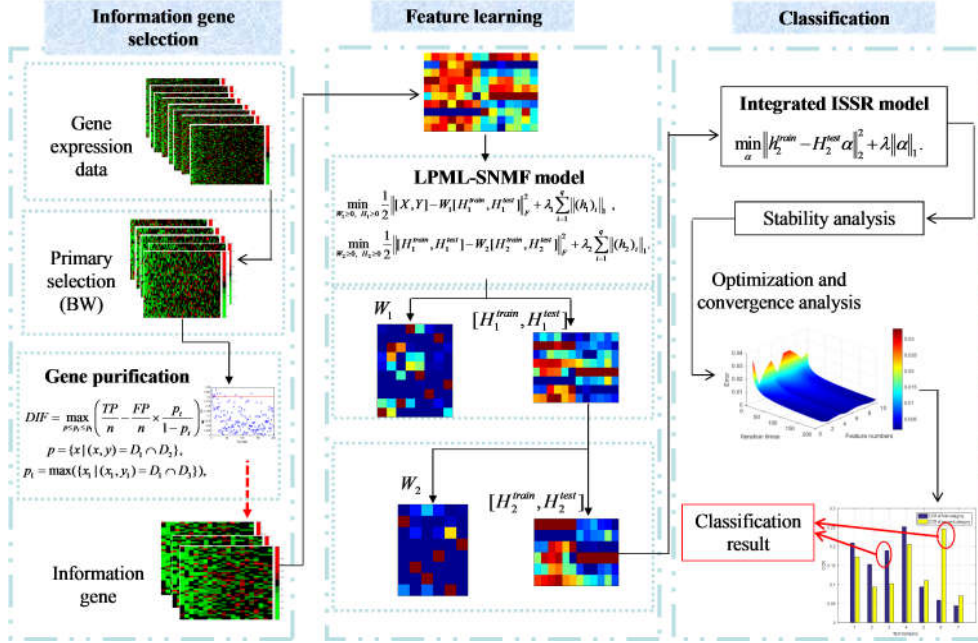

Fig. 1. Framework of the proposed tumor classification.

(1) An ISSRC model is constructed for alleviating the problems by insufficient training samples. The ISSRC model fully explores information embedded in existing available samples, especially test samples. The representation ability and classification stability of the ISSRC is similar to PFSRC [14] and superior to SRC [7], which relies on a large number of training samples.

(2) A DIF-based gene selection method is proposed to improve the representation ability of the ISSR model to small sample problem. Compared to existing gene selection methods, the proposed DIF-based technique is established for the first time by incorporating clinical misdiagnosis rate into gene selection.

(3) A layer-wise pre-training multi-layer sparse NMF (LPML-SNMF)-based feature learning method is proposed to further enhance the representation ability and classification stability of the ISSRC model, especially for unbalanced classification problem. The advantage of LPML-SNMF method is that it combines complementary strengths from NMF [24] and deep learning [33]. The hierarchical strategy enhances the representation learning ability of NMF by exploring the essential information contained in existing available training and test samples. The layer-wise pre-training strategy enhances the stability NMF by alleviating its sensitivity to iteration initials.

(4) The ISSRC combined with DIF-based gene selection and LPML-SNMF-based feature learning is called the integrated ISSRC, whose stability, optimization and the corresponding convergence are analyzed.

(5) The performance of the proposed integrated-based tumor classification framework is fully verified on six microarray gene expression datasets, which contain three stages of early diagnosis, tumor type recognition and postoperative metastasis.

The remainder of this paper is organized as follows. The methodology is given in Section 2, which mainly includes the construction, optimization and convergence analysis of the integrated ISSRC model. Extensive experiments on six public tumor gene expression datasets will be shown in Section 3. Finally, conclusions will be drawn in Section 4.

## 2. The integrated ISSR-based tumor classification

The integrated ISSRC model is proposed for microarray gene expression data-based tumor classification, and then the stability analysis, optimization and convergence analysis are given.

### 2.1 Construction of the Integrated ISSRC model

#### 2.1.1 ISSRC model

Firstly, an ISSRC model is proposed and its representation ability and classification stability will be analyzed. Suppose $X = [x_1, \cdots, x_{s_1}, \cdots, x_{s_c}] \in R^{d \times s_c}$ is a training sample set, $X_j = [x_{s_{j-1}+1}, \cdots, x_{s_j}] \in R^{d \times (s_j - s_{j-1})}$ are the $j$-th category samples, where $j = 1, \cdots, c$ is the number of category. $Y = [y_1, \cdots, y_k] \in R^{d \times k}$ is a test sample set. In SRC [7], each test sample $y_l$ can be linearly represented by the training sample set $X$. Without causing confusion, the corresponding projection way and representation space of SRC are called positive projection and positive space. PFSRC [14], by contrast, represents each training sample $x_i$ by its corresponding pseudo-full-space $V_i = \{X, Y\} - \{x_i\}$, $i = 1, \cdots, s_c$, where the projection way is inverse to SRC and called inverse projection. It is worth noting that the PFSRC aims to explore complementary information contained in available face samples. However, there is no such obvious complementarity between gene data, and there are few effective labeled training samples. To tackle this problem, an inverse space representation is proposed. In a sense, inverse space is a special case of pseudo-full-space.

**Definition 1 (Inverse space representation)** Suppose $Y$ is a test sample space, $x_i \in X$, $i = 1, \cdots, s_c$ are training samples. The inverse space representation means each training sample $x_i$ is represented by $Y$.

$$x_i = \alpha_{i,1} y_1 + \cdots + \alpha_{i,l} y_l + \cdots + \alpha_{i,k} y_k = \sum_{l=1}^{k} \alpha_{i,l} y_l = Y\alpha_i, \qquad (1)$$

where $\alpha_i = [\alpha_{i,1}, \cdots \alpha_{i,l}, \cdots, \alpha_{i,k}]^T$ is the representation coefficients of inverse space representation. The corresponding optimization problem can be written as,

$$\min_{\alpha_i} \|x_i - Y\alpha_i\|_2^2.$$

Considering there is an obvious sparse characteristic in microarray gene expression data, the sparsity constraint can be introduced into the inverse space representation and called the inverse space sparse representation (ISSR).

$$\min_{\alpha_i} \|x_i - Y\alpha_i\|_2^2 + \lambda \|\alpha_i\|_1, \qquad (2)$$

where $\lambda$ is regularization parameter, and $\alpha_i$ is the representation coefficient vector of $x_i$.

Similar to PFSRC [14], the category contribution rate (CCR) can be introduced to complete the classification. A test sample $y_l$ is classified into the category with the maximal CCR. It has been demonstrated that the PFSRC is more stable and effective than standard SRC, especially when there's a small number of training samples. Obviously, the ISSRC model inherits the advantages of the PFSRC in terms of representation ability and classification stability.

However, microarray gene expression data have the characteristics of small samples and high redundancy. How to further improve the representation stability and stability of the ISSRC model is interesting and necessary, especially there are a small number of training samples.

**2.1.2 DIF-based gene selection**

For further enhance the representation ability of the ISSRC model to small sample problem, a simple but effective quantitative index named DIF is established to select the small subset of information genes.

DCA [23] is a way of evaluating treatment plans by maximizing the clinic NB of profit minuses harm. As shown in Fig. 2, DCA evaluates a treatment plan by risk (of illness) – (clinic) NB ratio. The horizontal axis indicates when the risk of illness reaches a certain probability, the patient is considered to be positive and treatment is adopted. As shown in Fig. 2, the vertical axis indicates, after taking treatment, the corresponding NB of profit minus harm. The higher the NB is, the better the treatment plan is. It seems to be valuable to use DCA for predicting the usefulness of each gene. Let TP, TN, FP and FN denote the numbers of true-positive, true-negative, false-positive and false-negative of the patients. Suppose all the patients are negative, the NB is denoted as $D_1: NB = 0$ (blue dotted line in Fig.2), which is just the horizontal axis. Suppose all the patients are positive, the corresponding NB is denoted as, $D_2: NB = \frac{P}{n} - \frac{n-P}{n} \times \frac{p_t}{1-p_t}$ (black dotted line in Fig.2, where $p_t$ is a threshold probability. $n$ is the total number of patients and the prevalence ($p$) is just the intersection of $D_1$ and $D_2$.

In fact, the patients usually contain both TP and FP, so $D_2$ can be rewritten more generally as,

$$D_3: NB = \frac{TP}{n} - \frac{FP}{n} \times \frac{p_t}{1-p_t}. \tag{3}$$

It is worth noting that, for curve $D_3$ (green line in Fig.2), TP detection rate increases or FP decreases when NB is maximized. This means that misdiagnosis rate is reduced, which is exactly what clinical concerns and needs. Therefore, the treatment plan with the maximum NB will be adopted to obtain the lowest misdiagnosis rate.

On the other hand, [23] demonstrates that the effectiveness of the treatment in the areas among the three curves, $D_1$, $D_2$ and $D_3$, is valuable. Based on the characteristics of DCA, it is believed that DCA is suitable for predicting the usefulness of genes. The main idea of this lies in selecting information genes that can lead to the lowest misdiagnosis rate for clinical diagnosis. The higher the NB is, the lower the clinical misdiagnosis rate is, and the better the gene is. For convenience, a statistics index is defined to select information genes.

**Definition 2 (Decision information factor, DIF)** Suppose the threshold probability $p_t$ varies in the valid probability interval $[p, p_1]$, which is the intersection abscissa range of $D_1$, $D_2$ and $D_3$. Each treatment plan corresponds to a curve $D_3$ and the best one is just with the maximal value of NB. Similarly, each gene corresponds to a curve $D_3$ and the curve with the maximum NB will be focused. The point corresponding to the maximum NB is defined as a DIF index of a gene.

$$DIF = \max_{p \leq p_t \leq p_1} \left( \frac{TP}{n} - \frac{FP}{n} \times \frac{p_t}{1-p_t} \right),$$
$$p = \{x \mid (x, y) = D_1 \cap D_2\},$$
$$p_1 = \max(\{x_1 \mid (x_1, y_1) = D_1 \cap D_3\}),$$

where $DIF \in [0,1]$. Our purpose is to find information gene bringing the largest NB. The bigger the DIF value of a gene is, the higher benefit of the gene is to clinical diagnosis, and the better the gene is for classification.

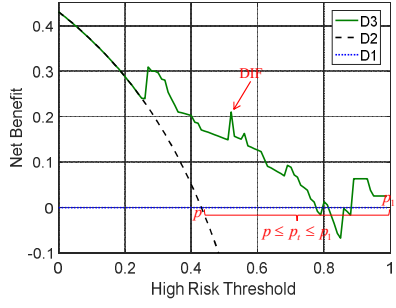

Fig.2. Construction of DIF index.

In short, the DIF-based gene selection method combined with the clinical misdiagnosis rate can simply and effectively select a small number of information genes, which are more likely to be used for tumor classification. The proposed DIF-based gene selection further enhances the representation ability of ISSRC model rather than original gene data.

**2.1.3 LPML-SNMF-based feature learning**

Based on the subset of information genes selected by DIF, a feature learning method named LPML-SNMF is proposed to further enhance the representation ability and classification stability of the ISSRC model. The superiority of the LPML-SNMF is that it complements the advantages of NMF and deep learning.

*A) Hierarchical representation learning strategy*

A hierarchical strategy is introduced into NMF, whose feature representation learning ability can be enhanced by deeply exploring more essential features than original gene data.

Suppose $V \in R^{d \times q}$ is a non-negative matrix, which is decomposed into two non-negative matrices $W$ and $H$, $V \approx WH$. The object function to be optimized is as follows,

$$\min_{W,H} \frac{1}{2} \|V - WH\|_F^2, \ s.t. \ W \geq 0, \ H \geq 0,$$

where $W \in R^{d \times r}$ is basis matrix and $H \in R^{r \times q}$ is coefficients matrix. Each column of $H$ is an encoding correspondence with $V$. The rank $r$ of the factorization is generally chosen so that $(d+q)r < d \times q$. Hoyer [32] proposed the SNMF, which added sparse regularization constraints to $H$. The corresponding objective function is revised as,

$$\min_{W,H} \frac{1}{2} \|V - WH\|_F^2 + \lambda_1 \sum_{i=1}^{q} \|h_i\|_1, \ s.t. \ W \geq 0, \ H \geq 0, \tag{4}$$

where $\lambda_1 > 0$ is a regularization parameter and $H = [h_1, h_2, \cdots, h_q]$, $h_i \in H$, $i = 1, \cdots, q$.

Motivated by deep learning, the hierarchical representation learning strategy is introduced and a multi-layer

SNMF (ML-SNMF) is performed on $V$,

$$\min_{W_1,\cdots,W_L,H_L} \frac{1}{2}\|V - W_1\cdots W_L H_L\|_F^2 + \lambda_1 \sum_{i=1}^{q}\|W_2\cdots W_L(h_L)_i\|_1 + \cdots + \lambda_L \sum_{i=1}^{q}\|(h_L)_i\|_1, \quad (5)$$
$$s.t. \ W_1 \geq 0, \cdots, W_L \geq 0, H_L \geq 0.$$

By comparing model (5) with model (4), one can notice that the representation ability is indeed enhanced by the deep representation learning.

*B) Layer-wise pre-training strategy*

A critical problem in NMF is that the results are heavily influenced by initial values. Layer-wise pre-training strategy can be introduced to mitigate the sensitivity of NMF to initial values and enhance its stability.

The model is based on the fact that the optimal output of the first layer is as the input of the second layer, and so on. Suppose the decomposition level is $L$, compared with the model (5), the model is as follows,

$$\min_{W_l,H_l} \frac{1}{2}\|H_{l-1} - W_l H_l\|_F^2 + \lambda_l \sum_{i=1}^{q}\|(h_l)_i\|_1, \quad s.t. \ W_l \geq 0, \ H_l \geq 0, \ l = 1,2,\cdots,L, \quad (6)$$

where the initial matrix $H_0$ represents $V$. By comparing model (6) with model (5), one can notice that the initialization effect of the NMF-based model can be alleviated to some extend by hierarchical representation learning and layer-wise pre-training strategy.

Similar to classical machine learning methods, NMF can be done on training samples. Suppose the training sample set $X$ is a non-negative matrix, which is decomposed into the corresponding non-negative basis matrix $W^{train} \in R^{d \times r}$ and coefficients matrix $H^{train} \in R^{r \times s_c}$. The model is as follows,

$$\min_{W_l^{train},H_l^{train}} \frac{1}{2}\|H_{l-1}^{train} - W_l^{train} H_l^{train}\|_F^2 + \lambda_l \sum_{i=1}^{s_c}\|(h_l^{train})_i\|_1, \quad s.t. \ W_l^{train} \geq 0, \ H_l^{train} \geq 0, \ l = 1,2,\cdots,L, \quad (7)$$

where the initial matrix $H_0^{train}$ represents training sample set $X$.

*C) LPML-SNMF model*

By combing the hierarchical and layer-wise pre-training strategies, a LPML-SNMF is constructed to explore available information embedded in the existing available training and test samples, especially when the training samples are small.

An observation shows that Eq. (7) depends heavily on the training samples, while gene-based tumor classification is a typical small sample problem. It is worth noting that there are usually a lot of unlabeled test samples that are not being used. NMF, however, has exactly the advantage of paying attention to category information. That is, NMF can make comprehensive use of training and test samples simultaneously. Therefore, the unlabeled test samples can be introduced into model (7) to improve the representation ability and stability of the model.

Suppose $V = [X,Y] \in R^{d \times (s_c + k)}$ is a collection of training samples and test samples after gene selection, where $q = s_c + k$. Compared with the model (7), the LPML-SNMF is as follows,

$$\min_{W_l, H_l} \frac{1}{2} \left\| [H_{l-1}^{train}, H_{l-1}^{test}] - W_l [H_l^{train}, H_l^{test}] \right\|_F^2 + \lambda_1 \sum_{i=1}^{q} \left\| (h_l)_i \right\|_1, \ s.t. \ W_l \geq 0, \ H_l \geq 0, \quad (8)$$

where the initial matrix $[H_0^{train}, H_0^{test}]$ represents the sample set $[X,Y]$, $W_l \in R^{d \times r}$ is basis matrix, $H_l = [H_l^{train}, H_l^{test}] = [(h_l)_1, \cdots, (h_l)_q] \in R^{r \times q}$ is coefficient matrix, , $H_l^{train} \in R^{r \times s_c}$ and $H_l^{test} \in R^{r \times k}$ are the $l$-th level coefficient matrices of training and test samples, respectively, $r_l$ is the $l$-th level rank of the matrix after feature learning, where $r_0$ represents $d$ and $r_l \ll \min\{r_{l-1}, q\}$. The corresponding improved NMF is called LPML-SNMF.

By comparing model (8) with model (7), one can notice that the LPML-SNMF integrates both training and test samples. The addition of the test samples makes the model can reflect internal essential information in test samples. Therefore, the LPML-SNMF model is more stable and more conductive for classification. See Subsection 3.4.2 for detailed experiments.

Taking two-layer model as an example, the LPML-SNMF model can be written as follows,

$$\min_{W_1, H_1} \frac{1}{2} \left\| [X,Y] - W_1 [H_1^{train}, H_1^{test}] \right\|_F^2 + \lambda_1 \sum_{i=1}^{q} \left\| (h_1)_i \right\|_1, \ s.t. \ W_1 \geq 0, \ H_1 \geq 0, \quad (9a)$$

$$\min_{W_2, H_2} \frac{1}{2} \left\| [H_1^{train}, H_1^{test}] - W_2 [H_2^{train}, H_2^{test}] \right\|_F^2 + \lambda_2 \sum_{i=1}^{q} \left\| (h_2)_i \right\|_1, \ s.t. \ W_2 \geq 0, \ H_2 \geq 0, \quad (9b)$$

where $W_1 \in R^{d \times r_1}$, $W_2 \in R^{r_1 \times r_2}$, and $(h_2)_i \in R^{r_2 \times 1}, i = 1, \cdots, q$ represent the second level coefficients corresponding to the $i$-th sample $v_i$, $H_2 = [H_2^{train}, H_2^{test}] = [(h_2)_1, \cdots, (h_2)_q] \in R^{r_2 \times q}$.

From the optimization point of view, each layer of LPML-SNMF model is similar to SNMF [32] and the variables are alternately iterated by gradient descent method. When the feature learning model is optimized by layer-wise pre-training technique, that is, the obtainable optimal solution of the previous layer is regarded as the input of the latter layer. The specific optimization process can see Appendix A. The comparison of LPML-SNMF and other improved NMF methods [31] is given in Subsection 3.4.2.

### 2.1.4 The integrated ISSRC model

*A) Construction of the integrated ISSRC model*

Based on the above proposed DIF-based gene selection and LPML-SNMF-based feature learning, an integrated ISSRC model is formed with more representation ability and classification stability.

Suppose the training feature sets $H_2^{train}$ and the test feature set $H_2^{test}$ obtained by the two-level LPML-SNMF (9), the integrated ISSR is as follows.

$$(h_2^{train})_i = \alpha_{i,1}(h_2^{test})_1 + \cdots + \alpha_{i,l}(h_2^{test})_l + \cdots + \alpha_{i,k}(h_2^{test})_k = \sum_{l=1}^{k} \alpha_{i,l}(h_2^{train})_l = H_2^{test}\alpha_i, \tag{10}$$

where $\alpha_i = [\alpha_{i,1}, \cdots \alpha_{i,l}, \cdots, \alpha_{i,k}]^T$ is the representation coefficients.

By comparing Eqs. (10) and (1), one can observe that the differences between the integrated ISSR and ISSR are representation space, an intuitive example is given in Fig. 3. Comparing Figs. 3 (a) and (b), it is easy to notice that the integrated ISSR focuses on the deeper and more essential characteristics contained in data, rather than the ISSR addresses the original data. The feature representation way makes the integrated ISSR less sensitive to the original samples than that of the ISSR, whether it's small sample or category-imbalance. As a result, the integrated ISSR is more stable and effective than ISSR.

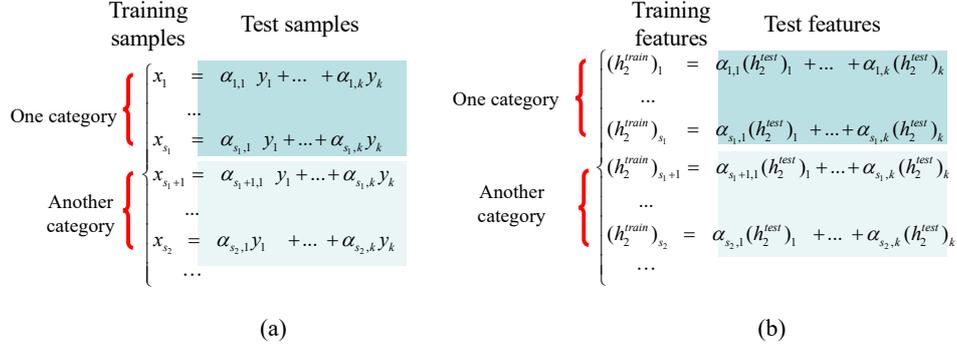

Fig.3. Comparison of different representation ways. (a) ISSR, (b) integrated ISSR.

For any $h_2^{train} \in H_2^{train}$, the integrated ISSR model represents $h_2^{train}$ by $H_2^{test}$,

$$\min_{\alpha} \left\| h_2^{train} - H_2^{test}\alpha \right\|_2^2 + \lambda \|\alpha\|_1, \tag{11}$$

where $\lambda > 0$ is a regularization parameter, and $\alpha$ is the representation coefficient vector of $h_2^{train}$.

The integrated ISSRC is constructed by the integrated ISSR model and the corresponding classification criterion, CCR, which is similar to [14].

*B) Stability analysis of integrated ISSRC model*

The stability analysis of the integrated ISSRC model and the corresponding stability theorem is given below.

**Theorem 2.1 (Classification stability of the integrated ISSRC)** Suppose $(h_2^{train})_i$ and $(h_2^{train})_j$ are the $i$-th and $j$-th training samples features, and the relationship $(h_2^{train})_i$ and $(h_2^{train})_j$ is $(h_2^{train})_j = (h_2^{train})_i + \Delta\left((h_2^{train})_i\right)$, where $\Delta\left((h_2^{train})_i\right)$ is a disturbance of $(h_2^{train})_i$. Based on the test samples features $H_2^{test}$, the inverse space representations of $(h_2^{train})_i, (h_2^{train})_j$ are as follows: $(h_2^{train})_i = H_2^{test}\alpha_i$, $(h_2^{train})_j = H_2^{test} \alpha_j$, where $\alpha_i$ and $\alpha_j$ are representation coefficients, respectively. Let $\Delta(H_2^{test})$ represents the disturbance corresponding to $\Delta((h_2^{train})_i)$. If

$$\varepsilon = \max \left\{ \frac{\|\Delta((h_2^{train})_i)\|_2}{\|(h_2^{train})_i\|_2}, \frac{\|\Delta(H_2^{test})\|_2}{\|H_2^{test}\|_2} \right\} \leq \frac{\varphi_k(H_2^{test})}{\varphi_1(H_2^{test})},$$

and $\sin(\theta) = \rho_{LS} / \|(h_2^{train})_i\|_2 \neq 1$, where $\rho_{LS} = \|H_2^{test} \alpha_{LS_i} - (h_2^{train})_i\|_2$, $\alpha_{LS_i} = \arg\min_{\alpha_i} \|(h_2^{train})_i - H_2^{test} \alpha_i\|_2$, then

$$\frac{\|\alpha_j - \alpha_i\|_2}{\|\alpha_i\|_2} \leq \varepsilon \left\{ \frac{2\kappa_2(H_2^{test})}{\cos(\theta)} + \tan(\theta) \kappa_2(H_2^{test})^2 \right\} + O(\varepsilon^2). \tag{12}$$

where $\kappa_2(H_2^{test})$ ($\kappa_2(H_2^{test}) = \|H_2^{test}\|_2 \cdot \|((H_2^{test})^T H_2^{test})^{-1}(H_2^{test})^T\|_2$, $\kappa_2(H_2^{test})^2 = \|H_2^{test}\|_2^2 \cdot \|((H_2^{test})^T H_2^{test})^{-1}\|_2$) is the $l_2$-norm conditional number of $H_2^{test}$, and $\theta$ is angle between $(h_2^{train})_i$ and its projection vector on $H_2^{test}$.

The conclusion indicates that the distance between $\alpha_i$ and $\alpha_j$ is very small when $(h_2^{train})_i$ is similar to $(h_2^{train})_j$ (in other words, $H_2^{test}$ has a small disturbance $\Delta(H_2^{test})$). From Eq. (12), one can see that coefficients are more sensitive to a small disturbance $\Delta$ than that of reconstruction error. Because, for nonzero residual problems, it is the square of the condition number that measures the sensitivity of coefficients. Moreover, it is worth noting that we focus on the column coefficient vector $\alpha_{1,1}, \alpha_{1,2}, \cdots, \alpha_{s_c,1}$ before each test sample when we calculate the CCR similar to [14]. The difference lies in the representation coefficients $\alpha$ of different representation spaces. However, it has been demonstrated that disturbance will affect row coefficients rather than column coefficients. Moreover, the effect on column coefficients is a positive impact when CCRs of different categories are calculated. Please see Appendix B for the detailed proof of the classification stability Theorem.

## 2.2 Optimization of the integrated ISSR model by GsADMM

The integrated ISSRC model can be optimized by GsADMM [39], which has a smaller convergence error and a faster convergence speed than the classic ADMM algorithm.

The integrated ISSRC model in Eq. (11) can be rewritten as

$$\min_{\alpha,b} \|h_2^{train} - H_2^{test} \alpha\|_2^2 + \lambda \|b\|_1 \quad s.t. \quad \alpha - b = 0. \tag{13}$$

For $h_2^{train} \in R^{r_2 \times 1}$, $H_2^{test} \in R^{r_2 \times k}$, the augmented Lagrangian function of (13) is defined as,

$$L_\sigma(\alpha, b; \eta) = \|h_2^{train} - H_2^{test} \alpha\|_2^2 + \lambda \|b\|_1 + \langle \eta, \alpha - b \rangle + \frac{\sigma}{2} \|\alpha - b\|_2^2, \tag{14}$$

Let $\sigma > 0$ be the penalty parameter, and $\eta \in R^{k \times 1}$ be the Lagrange multiplier, $\langle \cdot, \cdot \rangle$ denotes the inner product.

The GsADMM scheme takes the following form

$$\begin{cases} \alpha^k = \arg\min_{\alpha} L_\sigma(\alpha,\tilde{b}^k;\tilde{\eta}^k) + \frac{1}{2}\|\alpha - \tilde{\alpha}^k\|_K^2, & (a) \\ \eta^k = \tilde{\eta}^k + \sigma(\alpha^k - \tilde{b}^k), & (b) \\ b^k = \arg\min_{b} L_\sigma(\alpha^k,b;\eta^k) + \frac{1}{2}\|b - \tilde{b}^k\|_T^2, & (c) \\ \tilde{w}^{k+1} = \tilde{w}^k + \rho(w^k - \tilde{w}^k), & (d) \end{cases} \quad (15)$$

where $w^k = (\alpha^k, b^k, \eta^k)$, $\alpha^{-1} = \tilde{\alpha}^0$, $b^{-1} = \tilde{b}^0$, $K: R^{k\times 1} \to R^{k\times 1}$ and $T: R^{k\times 1} \to R^{k\times 1}$ are two semi-proximal matrixes. A more natural choice of the semi-proximal terms is to add $\frac{1}{2}\|\alpha - \tilde{\alpha}^k\|_K$ and $\frac{1}{2}\|b - \tilde{b}^k\|_T$ to the sub-problems for computing the values $\alpha^k$ and $b^k$. For the sake of generality and numerical convenience, the latter variant with only semi-proximal terms is considered. The most adopted values of the variables are used in the proximal terms.

Please see Appendix C for detailed optimization process of the integrated ISSRC model.

## 2.3 Convergence analysis

Convergence analysis is crucial to optimization. The convergence theorem and the corresponding lemmas are given below. In order to prove the theorem, Karush-Kuhn-Tucker (KKT) for model (13) is given first.

Let $f(\alpha) = \|h_2^{train} - H_2^{test}\alpha\|_2^2$, $g(b) = \lambda\|b\|_1$, a vector $(\tilde{\eta}, \tilde{\alpha}, \tilde{b})$ is a saddle point to the Lagrangian function if it is a solution to the following KKT system

$$\eta \in \partial f(\alpha), \quad -\eta \in \partial g(b), \text{ and } \alpha - b = 0. \quad (16)$$

Next, let $(\bar{\eta}, \bar{\alpha}, \bar{b})$ be an arbitrary solution to the KKT system (16). For any $(\eta, \alpha, b)$, we denote $\eta_e = \eta - \bar{\eta}$, $\alpha_e = \alpha - \bar{\alpha}$ and $b_e = b - \bar{b}$. In order to give the convergence theorem of integrated ISSR model based on GsADMM optimization, two lemmas are given below.

**Lemma 2.1** Let $(\bar{\eta}, \bar{\alpha}, \bar{b})$ be a solution to the KKT system (16) and $(\eta^k, \alpha^k, b^k)$ be the sequence generated by Eq. (15). For any $k \geq 0$, the following equations hold.

$$\langle \eta^{k+1} - \eta^k, \alpha^{k+1} - \alpha^k \rangle - \frac{\rho}{2}\|\alpha^{k+1} - \tilde{\alpha}^{k+1}\|_K^2 + \frac{\rho}{2}\|\alpha^k - \tilde{\alpha}^k\|_K^2 \geq \|\alpha^{k+1} - \alpha^k\|_{\Sigma_f}^2, \quad (17)$$

and

$$\frac{\sigma\rho}{2}\|\alpha_e^{k+1} - b_e^k\|^2 + \langle \eta_e^{k+1} + \sigma(1-\rho)\alpha_e^{k+1}, \alpha_e^{k+1} - b_e^k \rangle = -\frac{1}{2\sigma\rho}[\|\eta_e^{k+1} + \sigma(1-\rho)\alpha_e^{k+1}\|^2 - \|\eta_e^k + \sigma(1-\rho)\alpha_e^k\|^2]. \quad (18)$$

**Lemma 2.2** Assume Eq. (15) holds and the sequence $\{(\eta, \alpha, b)\}$ is generated by Eq. (15). Then for any

$k \geq 0$, one can get

$$\phi_k - \phi_{k+1} \geq 2\|\alpha_e^k\|_{\Sigma_f}^2 + 2\|b_e^k\|_{\Sigma_g}^2 + \sigma(2-\rho)\|\alpha_e^{k+1} - b_e^k\|^2 + (2-\rho)\|\tilde{b}^k - b^k\|_T^2 + (2-\rho)\|\tilde{\alpha}^{k+1} - \alpha^{k+1}\|_K^2 \\ + \sigma\rho^{-1}(2-\rho)^2\|\alpha_e^k - \alpha_e^{k+1}\|^2 + 2(2-\rho)\rho^{-1}\|\alpha^{k+1} - \alpha^k\|_{\Sigma_f}^2. \tag{19}$$

Now, we are ready to establish the global convergence of Eq. (15).

**Theorem 2.2** Suppose there exists a vector $(\tilde{\alpha}, \tilde{\eta}, \tilde{b})$ satisfying the KKT system. Let $\{(\eta^k, \alpha^k, b^k)\}$ be the sequence generated by Eq. (15). Then the whole sequence $\{(\eta^k, \alpha^k, b^k)\}$ converges to a solution to the KKT system.

Theorem 2.2 enlightens that if the solution of the model exists, the iterative solution satisfies the constraint condition. Furthermore, if the solution is unique, the iterative solution of each single variable converges to the real solution. Please see Appendix D for the detailed proof of the convergence Theorem 2.2. Convergence analyses are verified in the experimental subsection 3.4.3 B).

For convenience, the integrated ISSR-based tumor classification is called the integrated ISSRC. The corresponding algorithm is given as follows.

---

**Algorithm 1: The integrated ISSRC algorithm**

**Input:** Training sample set $X = [x_1, \cdots, x_{s_c}]$, training label set $L = [l_1, l_2 \cdots, l_{s_c}]$ and test sample set $Y = [y_1, y_2, \cdots, y_k]$.

**Gene selection step**
1) DIF-based gene selection is based on BW-based gene pre-selection.
2) By $DIF = \max_{p \leq p_t \leq p_1}\left(\frac{TP}{n} - \frac{FP}{n} \times \frac{p_t}{1-p_t}\right)$, $p = \{x \mid (x,y) = D_1 \cap D_2\}$, $p_1 = \max(\{x_1 \mid (x_1, y_1) = D_1 \cap D_3\})$, every pre-selection gene DIF are obtained.
3) The DIF is sorted in descending order, and the genes corresponding to the first 10 DIFs are selected as the information gene subset.

**Feature representation learning step**
The information genes selected based on DIF importing LPML-SNMF model.
1) By model $\min_{W_1, H_1} \frac{1}{2}\|[X,Y] - W_1[H_1^{train}, H_1^{test}]\|_F^2 + \lambda_1 \sum_{i=1}^q \|(h_1)_i\|_1$, s.t. $W_1 \geq 0$, $H_1 \geq 0$, the first layer of LPML-SNMF feature learning is realized.
2) By model $\min_{W_2, H_2} \frac{1}{2}\|[H_1^{train}, H_1^{test}] - W_2[H_2^{train}, H_2^{test}]\|_F^2 + \lambda_2 \sum_{i=1}^q \|(h_2)_i\|_1$, s.t. $W_2 \geq 0$, $H_2 \geq 0$, the second layer of LPML-SNMF feature learning is realized.

**Classification step**
Feature learning based on LPML-SNMF and classification based on integrated ISSRC.
1) For the training feature set $H_2^{train}$ and the test feature set $H_2^{test}$ obtained by the two-level LPML-SNMF. The integrated ISSR model is realized based on $\min_{\alpha} \|h_2^{train} - H_2^{test}\alpha\|_2^2 + \lambda\|\alpha\|_1$.
2) By subsection 2.2 and appendix C, for the optimization process, the projection coefficient matrix is obtained.
3) By $C_{j,l} = \frac{1}{s_j}\sum_i \left(\delta_j(\{|\alpha_{i,l}|\})/\sum_i\{|\alpha_{i,l}|\}\right)$, $i = 1, \cdots s_1, \cdots s_c$, the CCR matrix is obtained, relevancies between each test sample and all categories are obtained.

**Output:** Each test sample can be classified into the category with the maximal CCR.

---

## 3. Experiments and discussions

The performance of the proposed method will be demonstrated on three stage datasets: early diagnosis, tumor type recognition and postoperative metastasis. Firstly, early diagnosis is done on Colon dataset [40], and compared with other state-of-the-art SRC methods and the latest published classification results. Secondly, tumor type recognition is done on DLBCL [41] and Leukemia [42] datasets, and compared with other state-of-the-art SRC methods and the latest published classification results. Finally, postoperative metastasis is deeply analyzed on three Breast datasets [3], which is fully verified by verifying the performance of gene selection, feature learning and classification. Moreover, meaningful biological analysis of the selected pathogenic genes is made by enrichment analysis and survival curve analysis. Without loss of generality, the 10-fold cross-validation is used. All experiments have been carried out using MATLAB R2016a on a 3.30GHz machine with 4.00GB RAM and R-3.5.0.

### 3.1 Tumor datasets

The dataset of early diagnosis: Colon [40] is a binary category dataset, which consists of 40 tumor and 22 normal colon tissue samples. Each sample has 2000 genes.

The dataset of tumor type recognition: DLBCL [41] dataset consists of gene expression data of diffuse large B cell lymphoma, follicular lymphoma. There are 77 samples, each of which contains 5469 genes. Leukemia [42] dataset consists of gene expression data of acute myelogenous leukemia, acute lymphoblastic leukemia and mixed-lineage leukemia, including 72 samples. Each sample has 11225 genes.

The dataset of postoperative metastasis: Breast-2 is a dataset of the primary breast tumors of 25,000 genes from 117 young patients. In [3], 79 patients with ages under 55 with primary lymph node-negative breast tumor are selected for testing, where 34 from patients who have developed distant metastases within 5 years, and 45 from patients who are disease-free after a period of at least 5 years. From all patients, tumor sizes are under 5cm. The patients who have developed distant metastases within 5 years are recorded as tumor samples without any confusion, the patients who are continues to be disease-free after a period of at least 5 years are called a normal sample. Breast-2(77) is a subset of Breast-2 [3]. There are 44 developed distant metastases within 5 years and 33 remained to be metastases free for at least 5 years. Breast-2(97) is another subset of Breast-2. There are 97 lymph node-negative breast tumor patients. Among them, 46 developed distant metastases within 5 years and 51 are remained to be metastases free for at least 5 years.

## 3.2 Recognition of early diagnosis

In the same experimental environment and on the same dataset [40], the performance of the proposed integrated ISSRC method are compared with the latest published classification results [11-12, 20, 44-49], and some other state-of-the-art SRC methods [7, 14, 43]. Table 1 shows that the classification accuracies of our method are higher than those of in the nine latest published results. Especially, the classification accuracy of our method achieves 98.70% and is much higher than other methods. Table 2 gives the extensive experiment results conclude accuracy, sensitivity, specificity. One can observe that our method has a significant advantage than other approaches. All this suggests that our approach is effective in early identification (normal or tumor).

Table 1 Classification performance with the latest published results on Colon dataset

| Experiments | Methods | Accuracy (%) |
| --- | --- | --- |
| Deng et al.(2013) [48] | GRRF-RF | 82.50 |
| García et al. (2015) [45] | MLP-D | 83.74 |
| Dettling et al.(2004) [46] | BagBoost | 83.90 |
| Ruiz et al.(2006) [20] | BIRS+NB | 85.48 |
| Younsi et al.(2016) [49] | αRSE | 86.98 |
| Zheng et al.(2011) [12] | MSRC-SNMF | 90.32 |
| Gan et al. (2014) [11] | SRC-LatLRR | 90.32 |
| Gan et al.(2016) [47] | MRSRC-SVD | 90.32 |
| Liu et al.(2015) [44] | RPCA+LDA+SVM | 90.45 |
| **Our paper** | **Integrated ISSRC** | **98.70** |

Table 2 Classification results based on different methods on Colon dataset

| Methods | Accuracy (%) | Sensitivity (%) | Specificity (%) |
| --- | --- | --- | --- |
| SRC [7] | 89.28 | 90.00 | 88.33 |
| RRC_L1 [43] | 92.14 | 95.00 | 90.83 |
| RRC_L2 [43] | 90.71 | 92.50 | 90.83 |
| PFSRC [14] | 93.81 | 92.50 | 93.33 |
| **Integrated ISSRC** | **98.70** | **97.50** | **100** |

## 3.3 Recognition of tumor types

In the same experimental environment and on the same datasets [41, 42], the performance of the proposed integrated ISSRC are compared with the latest published classification results [11-12, 20, 45-48, 50-51], and some other state-of-the-art SRC methods [7, 14, 43]. Table 3 shows that, the classification accuracies of our method are higher than those of the latest published results except for Gan et al. [11]. From Table 4, one can observe that our method has a significant advantage in identifying different types of tumors.

Table 3 Classification performance with the latest published results on DLBCL and Leukemia datasets

| Experiments | Methods | Accuracy (%) |
|---|---|---|
| DLBCL dataset | | |
| García et al. (2015) [45] | MLP-D | 96.24 |
| Hong et al.(2009) [51] | Gene boosting-KNN | 97.20 |
| Zheng et al.(2011) [12] | MSRC-SNMF | 97.40 |
| Gan et al. (2014) [11] | SRC-LatLRR | 97.40 |
| **Our paper** | **Integrated ISSRC** | **97.50** |
| Leukemia dataset | | |
| Piao et al.(2012) [50] | ECBGS | 90.28 |
| Deng et al.(2013) [48] | GRRF-RF | 92.00 |
| Ruiz et al.(2006) [20] | BIRS+NB | 93.04 |
| Zheng et al.(2011) [12] | MSRC-NMF | 95.83 |
| Dettling et al.(2004) [46] | BagBoost | 95.92 |
| Gan et al.(2016) [47] | MRSRC-SVD | 97.22 |
| Gan et al. (2014) [11] | SRC-LatLRR | **98.61** |
| **Our paper** | **Integrated ISSRC** | **98.61** |

Table 4 Classification results based on different methods on DLBCL and Leukemia datasets

| Methods | Accuracy (%) | Sensitivity (%) | Specificity (%) |
|---|---|---|---|
| DLBCL dataset | | | |
| SRC [7] | 96.07 | 84.21 | 100 |
| RRC_L1 [43] | 96.25 | 94.74 | 96.55 |
| RRC_L2 [43] | 94.82 | 89.47 | 96.55 |
| PFSRC [14] | 93.57 | 94.74 | 93.10 |
| **Integrated ISSRC** | **97.50** | **100** | **96.55** |
| Leukemia dataset | | | |
| SRC [7] | 91.67 | 100 | 87.50 |
| RRC_L1 [43] | 93.06 | 95.83 | 91.67 |
| RRC_L2 [43] | 91.78 | 95.83 | 89.58 |
| PFSRC [14] | 94.28 | 95.83 | 93.75 |
| **Integrated ISSRC** | **98.61** | **100** | **97.92** |

### 3.4 Recognition of postoperative metastasis

From subsections 3.2 and 3.3, one can see that the proposed integrated ISSRC method achieves good recognition effects not only on early diagnosis dataset but also on tumor type recognition datasets. Surgery can remove the tumor to some extend, however, there is residual cancer, regional lymph node metastasis, or the presence of cancer emboli in the blood vessels, the risk of recurrence and metastasis is still very high. Residual cancer cells develop rapidly in patients with weak immunity and form new lesions. Therefore, it is necessary and important to identify the metastasis of cancer after surgery.

In this subsection, the effectiveness and stability of the proposed method are demonstrated through comprehensive and in-depth experiments on three breast tumor gene expression datasets. The experiments include the following aspects: (1) the proposed DIF-based gene selection compared with those of BW [17], signal noise ratio (SNR) [18] and the latest ROC gene selection method proposed in [21]; (2) the performance of LPML-SNMF is compared with ML-NMF [31], SNMF [32] and MI-SNMF; (3) classification performance is not only compared with those of the traditional classification methods NN [52], SVM [6], CRC [13] and SRC [7], but also compared with the latest SRC methods, such as PFSRC [14], RRC_L1 [43] and RRC_L2 [43]; (4) many kinds of measures are used to measure the performance of these methods, such as accuracy, sensitivity, positive predictive value, negative predictive value, error reduction rate (ERR) [53], ROC [22], DCA [23], heatmap, correlation coefficient (CC) and box plots; (5) the biological analysis of the selected information genes.

### 3.4.1 Performance of DIF-based gene selection

In this subsection, the effectiveness and efficiency of the proposed DIF-based information gene selection method is demonstrated. DIF-based gene selection is based on BW-based gene pre-selection. Here, 200 genes are pre-selected by BW.

Fig.4 shows that the DIF values of the 200 pre-selected genes. One can observe that most of the DIF values concentrate in the interval of $[0, 0.15]$, where the red line is the threshold for selecting genes. In this paper, the top-ranked 10 genes are selected that correspond to the maximum DIF values.

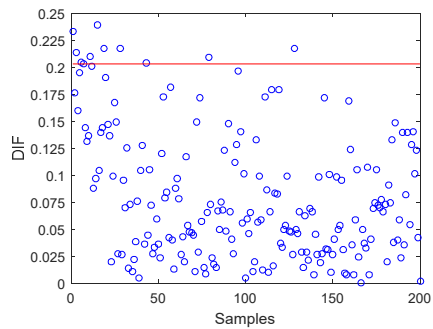

Fig. 4. Gene selection based on DIF. The blue circles are DIF values of the 200 genes selected by BW, the red line is the threshold for selecting the top-ranked 10 genes.

The performance of gene selection will be conducted using DCA and the principal component analysis (PCA). The DCA is adopted to furthermore demonstrate the performance of the top-10 information genes selected by DIF. DCA curves of the top-10 genes based on BW pre-selection (green curves) and those of the proposed DIF selection (red curves) are shown in Fig.5, where red curves are higher than green curves in the threshold interval. The higher the decision curve is, the greater the net benefit is, and the lower the clinical misdiagnosis rate of the classification is. Fig.6 gives 79 samples consisting of 34 tumor (red stars) and 45 normal (blue squares) using the top three principal components of 200 genes based on BW gene primary selection and 10 genes based on DIF, respectively. From Fig.6 (a) to (c), it is obvious that the normals and tumors become more and more distinguishable. Fig.5 and Fig.6 show that the superiority of applying the proposed DIF-based gene selection. However, it can be also seen form Fig. 6 (c) that there is still a degree of confusion that affects recognition.

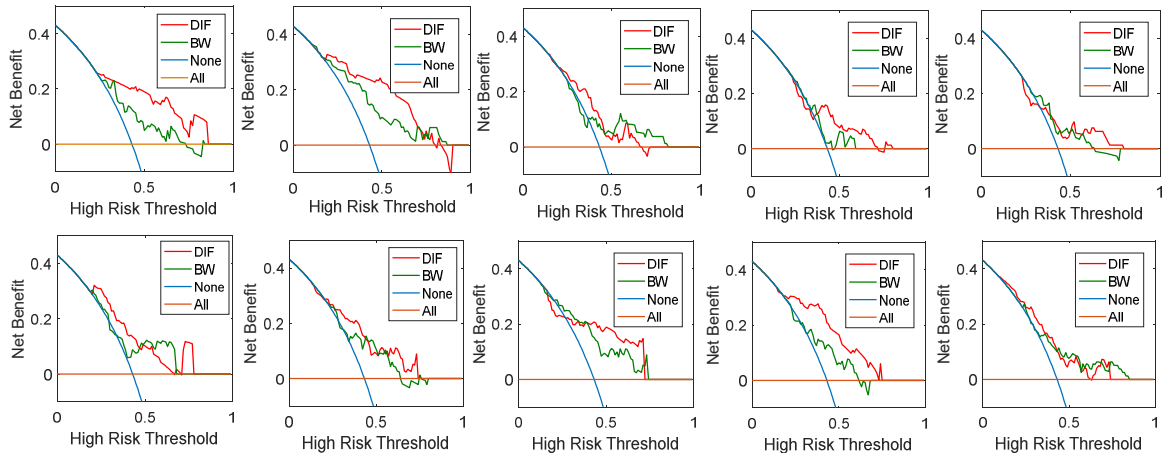

Fig. 5. Comparison of 10 genes selected based on DIF and BW on Breast-2 dataset.

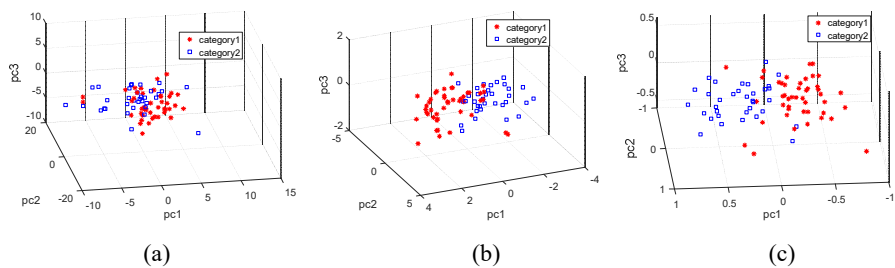

Fig. 6. Visualization of the first three principal components of PCA. Representation of all samples consisting of red stars and blue squares corresponds to 45 normals and 34 tumors on Breast-2 dataset. (a) original genes, (b) BW-based 200 genes; (b) DIF-based 10 genes.

For further accessing the performance of the DIF-based gene selection, experiments are conducted on Breast-2 dataset. Compared methods contain original gene data (no gene selection), BW [17], SNR [18], ROC [21] and DIF. The same classification method ISSRC is adopted. It can be seen from Table 5 that the classification performance of DIF is superior to all the other compared methods. In conclusion, DIF-based information genes selection has greatly improved the classification performance.

Table 5 Comparison of different gene selection methods.

| Methods | Original gene data | BW [17] | SNR [18] | ROC [21] | **DIF** |
|---|---|---|---|---|---|
| Accuracy (%) | 58.39 | 61.15 | 62.76 | 59.88 | **70.97** |

### 3.4.2 Performance of LPML-SNMF-based feature representation learning

In this subsection, the effectiveness and efficiency of the proposed LPML-SNMF-based feature learning method are demonstrated. Without causing confusion, $V$ represents the original information genes matrix, $H_1$ and $H_2$ represent the first and second layer feature matrix of LPML-SNMF, respectively. The decomposition dimensions corresponding to the first and second layer are $r_1 = 8$ and $r_2 = 6$ by experience and experiments.

### A) Performance of feature representation

The representation ability of the integrated ISSRC model will be verified. The comparison of representation coefficients before and after adding the test samples into the integrated ISSRC is shown in Fig. 7. The first 40 training samples are normals, and the last 30 are tumors. In Fig.7, the green curves denote the integrated ISSRC representation coefficient of the normals and the red curves expressed those of the tumors. The horizontal straight lines indicate the mean values of representation coefficients in the corresponding category. The difference between the two means of the LPML-SNMF model with test samples is much more obvious than that with training samples only. Therefore, one can conclusion that the LPML-SNMF model added the test samples can increase the representation of the integrated ISSR model.

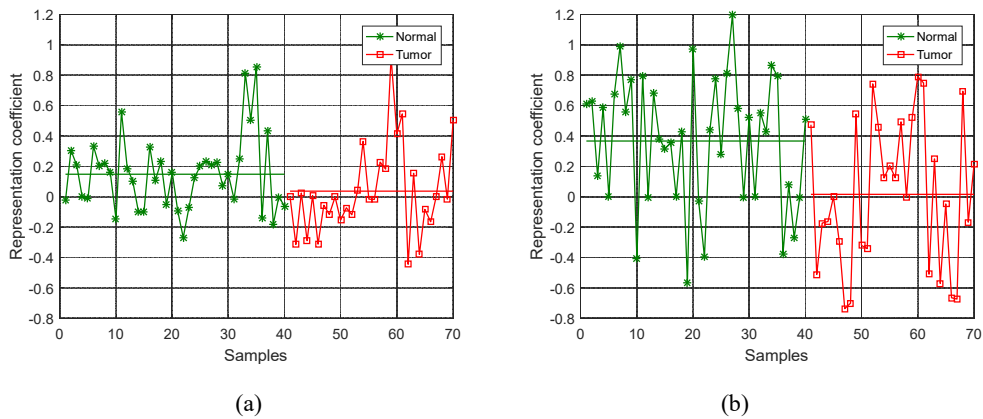

Fig.7. The integrated ISSR representation coefficient of different features. (a) the integrated ISSR representation coefficient of the feature obtained LPML-SNMF-0, (b) the integrated ISSR representation coefficient of the feature obtained LPML-SNMF-1.

For further accessing the performance of the test samples added for LPML-SNMF model, experiments are conducted on Breast-2 dataset. Without causing confusion, the LPML-SNMF model using only training samples is called LPML-SNMF-0, while the model using both training and test samples is called LPML-SNMF-1. Compared methods contain the first layer and the second layer of LPML-SNMF-0 and LPML-SNMF-1. The same classification method ISSRC is adopted. It can be seen from Table 6, in either case, the classification accuracies of the LPML-SNMF-1 is higher than that of the LPML-SNMF-0. The LPML-SNMF-1 model shows good and stable classification performance.

Table 6 Classification accuracies (%) of LPML-SNMF on Breast-2 dataset.

| Methods | The first layer | The second layer |
| --- | --- | --- |
| LPML-SNMF-0 | 74.35 | 88.59 |
| **LPML-SNMF-1** | **77.14** | **96.03** |

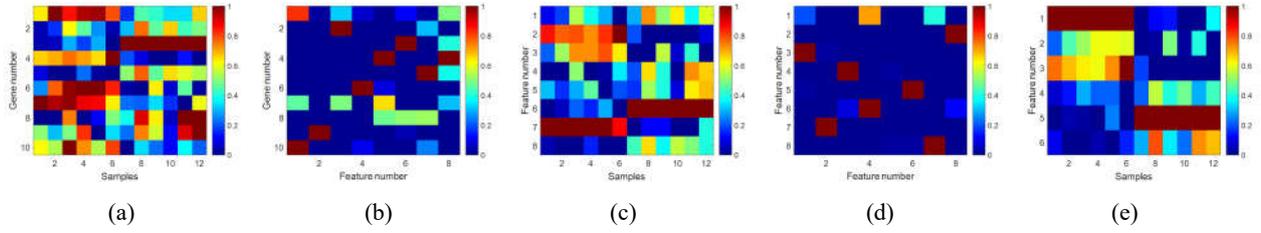

(a)                (b)                (c)                (d)                (e)

Fig. 8. LPML-SNMF feature learning. (a) Original matrix $V$ of all information genes, (b) and (d) are the first and second layer basis matrix $W_1$ and $W_2$, (c) and (e) are the first and second layer feature matrix $H_1$ and $H_2$.

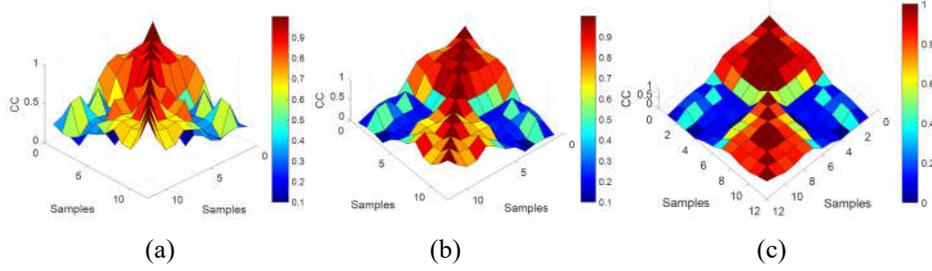

(a)                (b)                (c)

Fig. 9. The CC between samples as a three-dimensional heatmap. (a) the original matrix of all information genes $V$, (b) the first layer feature matrix $H_1$ of LPML-SNMF, (c) the second layer feature matrix $H_2$ of LPML-SNMF.

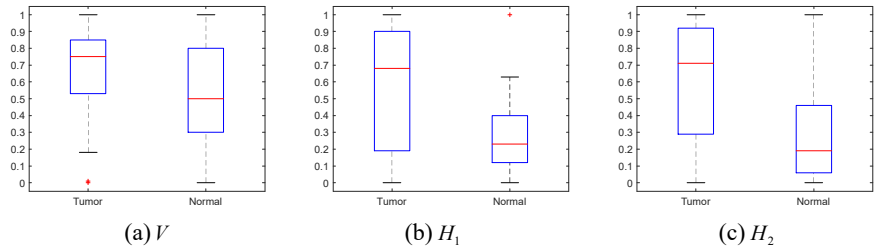

(a) $V$                (b) $H_1$                (c) $H_2$

Fig.10. LPML-SNMF features of each layer box plots

The performance of LPML-SNMF-based feature learning is analyzed by heatmap, correlation and box plots. Heatmap is an intuitive visualization method for analyzing the distribution of experimental data. Correlation analysis is used to analyze the correlation of the LPML-SNMF features obtained by each layer. Fig.8 shows the heatmap of LPML-SNMF-based feature learning, where blue to red colors represent low to high expression levels of genes or features. Fig.8 (a) represent the original information genes matrix $V$, Figs.8 (b) and (d) are the first and second layer basis matrices $W_1$ and $W_2$, and Figs.8 (c) and (e) are the first and second layer feature matrices $H_1$ and $H_2$. Fig.8 shows that the representation ability of features becomes stronger with feature learning. This is reflected in the more and more similar gene expression levels of the same category samples, and the more and more different gene expressions of difference categories samples. Fig.9 shows that the CC between samples as a three-dimensional heatmap, where blue to red color represents low to high expression levels of CCs. Fig.9 (a) represents the matrix of all information genes $V$, Figs.9 (b) and (c) represent the first and second layer feature matrix $H_1$ and $H_2$ of LPML-SNMF. From Fig.9 (a) to Fig. 9 (c), it can be seen that the one diagonal is getting red and the other diagonal is getting blue, which shows the relevancy of the same categories is increasing, and that of different categories is decreasing. Fig. 10 is the comparison of the box plot, one can see that the median of the two categories of samples is farther apart from each other with increasing the number of the decomposition layers. The expression level of tumor samples is getting lower and lower. Figs. 8, 9 and 10 show that the classification performance of features improves as the number of the LPML-SNMF decomposition layers increases.

**B) Regularization parameters analysis**

As shown in subsection 2.1.3, there are two regularization parameters, $\lambda_1$ and $\lambda_2$, which control the sparsity of matrix $H_1$ and $H_2$ in LPML-SNMF model (9a) and (9b). It is well known that these parameters have a great effect on the overall performance of the model. Therefore, the setting of this set of parameters is tested, and the appropriate values of $\lambda_1$, $\lambda_2$ are selected through the experimental results, in the case of the same initial value $W_1^0$, $W_2^0$, $H_1^0$, $H_2^0$. Fig. 11 shows that the regularization parameters $\lambda_1 = 0.2, \lambda_2 = 0.5$ are the best one corresponding to the classification results. As a result, this set of parameters is adopted.

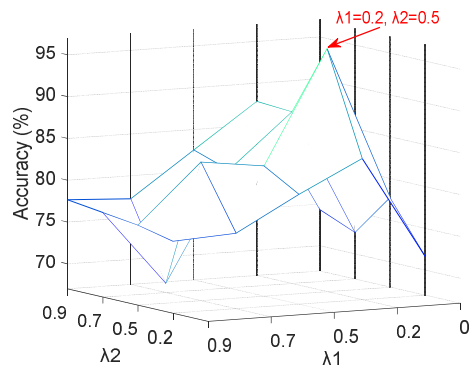

Fig.11. Accuracies based on ISSRC with different $\lambda_1$, $\lambda_2$ three-dimensional surface on Breast-2 dataset.

**C) Performance of classification**

In this subsection, the classification performance of the LPML-SNMF-based feature learning method is demonstrated. It is well known that postoperative metastasis is essential for breast tumor, and low missed diagnosis rate is needed for clinical use.

It can be seen from Table 7 that the classification performance of feature learning is better than those of raw information genes data before feature learning, and LPML-SNMF is superior to other feature learning methods under all gene selection methods. Table 8 gives the extensive experiments conclude accuracy, sensitivity, specificity, missed diagnosed, misdiagnosis, positive predictive value and negative predictive value. Table 8 shows that the LPML-SNMF has higher accuracy, sensitivity, specificity, positive predictive value and negative predictive value, and lower missed diagnosed, and misdiagnosis. All the results demonstrate the advantages of the LPML-SNMF-based feature learning method for tumor classification.

Table 7 Classification accuracies of feature learning based on different gene selection methods.

| Methods | Information genes | SNMF [32] | ML-NMF [31] | ML-SNMF | LPML-SNMF |
|---|---|---|---|---|---|
| BW [17] | 61.15 | 67.70 | 79.72 | 81.15 | 86.05 |
| SNR [18] | 62.76 | 65.12 | 76.89 | 78.611 | 86.51 |
| ROC [21] | 59.88 | 69.86 | 80.34 | 81.31 | 88.59 |
| **DIF** | <u>70.97</u> | <u>77.14</u> | <u>84.94</u> | <u>87.94</u> | <u>96.03</u> |

Table 8 Classification accuracies of feature learning based on different classification methods.

| Indexes | Methods | SNMF [32] | ML-NMF [31] | ML-SNMF | LPML-SNMF |
|---|---|---|---|---|---|
| Accuracy | **ISSRC** | 77.14 | 84.94 | 87.94 | <u>96.03</u> |
|  | SRC | 73.08 | 80.97 | 82.08 | 90.63 |
| Sensitivity | **ISSRC** | 77.50 | 84.50 | 87.00 | <u>97.50</u> |
|  | SRC | 75.00 | 84.50 | 86.50 | 93.00 |
| Specificity | **ISSRC** | 76.67 | 85.83 | 89.17 | <u>94.17</u> |
|  | SRC | 70.00 | 75.83 | 75.83 | 87.50 |
| Missed diagnosed | **ISSRC** | 22.50 | 15.50 | 13.00 | <u>2.50</u> |
|  | SRC | 25.00 | 15.50 | 13.50 | 7.00 |
| Misdiagnosis | **ISSRC** | 23.33 | 14.17 | 10.83 | <u>5.83</u> |
|  | SRC | 30.00 | 24.17 | 24.17 | 12.50 |
| Positive predictive value | **ISSRC** | 82.67 | 90.33 | 90.50 | <u>95.83</u> |
|  | SRC | 79.50 | 84.83 | 84.83 | 93.00 |
| Negative predictive value | **ISSRC** | 72.00 | 75.00 | 86.67 | <u>96.67</u> |
|  | SRC | 67.83 | 85.00 | 87.00 | 93.00 |

### 3.4.3 Performance of the integrated ISSRC

In this subsection, the performance of the integrated ISSRC is verified. Firstly, the feasibility of the integrated ISSRC model is verified. Secondly, the convergence of integrated ISSR model by GsADMM optimization is verified. Thirdly, it is verified that the integrated ISSRC model alleviates the classification unstable problem to some extent. Finally, classification performance of the integrated ISSRC is verified by comparing with some classical classification methods, nearest neighbor (NN) [52], SVM [6], CRC [13] and SRC [7], and some newly SRC improvement methods, PFSRC [14], RRC_L1 [43] and RRC_L2 [43]. Without loss of generality, all the classification results are all based on the DIF-based information selection and LPML-SNMF-based information feature learning.

### A) Feasibility analysis of the integrated ISSRC

In order to verify the feasibility of the integrated ISSRC, experiments are performed based on the same classification method ISSRC. The performance of the integrated ISSRC is compared with those of ISSRC (classification based on original gene data), DIF-based gene selection and ISSRC-based classification (no feature learning), LPML-SNMF-based feature learning and ISSRC-based classification (no gene selection). Table 9 shows that both DIF-based gene selection and LPML-SNMF-based feature learning can improve classification performance. The classification result of the integrated ISSRC is higher than that of single addition gene selection and feature learning. Hence, the following experiments are based on the integrated ISSRC model.

Table 9 Classification accuracies of different methods.

| Methods | ISSRC | DIF+ISSRC | LPML-SNMF+ISSRC | **Integrated ISSRC** |
|---|---|---|---|---|
| Accuracy (%) | 58.39 | 70.97 | 87.16 | **96.03** |

**B) Convergence analysis of the integrated ISSR model by GsADMM optimization**

In subsection 2.2, the integrated ISSR model is optimized by GsADMM. Here, the corresponding convergence is analyzed and compared with the classic ADMM. The convergence results are shown in Fig. 12.

Fig.12 (a) is the iteration error between exact and iterative solutions, Fig.12 (b) is the iteration error between the adjacent iterations. And Fig. 12 (c) gives the trend graph, which shows that the solution gradually becomes stable and converges to the numerical solution. It can be seen from Fig.12 that the iterative rate of GsADMM (red line) is faster than ADMM (blue line), while the convergence error of GsADMM is less than that of ADMM. Specifically speaking, the convergence error of ADMM is about 0.002, and iteration time is about 80s. The convergence error of GsADMM is about 0, and iteration convergence time is about 5s. Fig. 12 demonstrates that GsADMM is superior to the classic ADMM when solving the integrated ISSR model. Therefore, GsADMM is adopted in this paper.

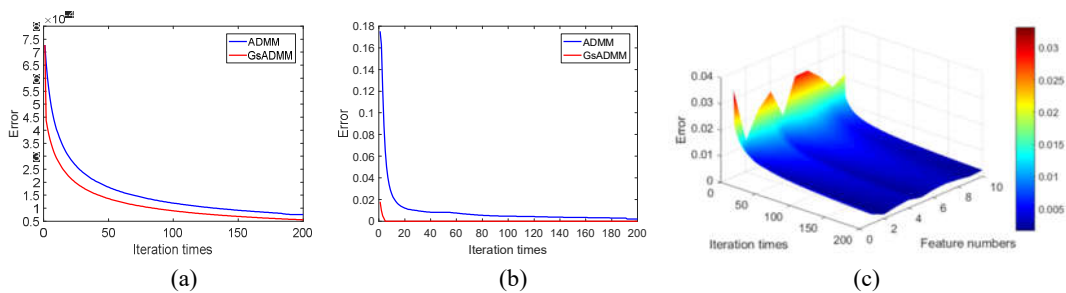

Fig. 12. Convergence analysis of the integrated ISSR model. (a) The iteration error between the exact and iterative solutions; (b) The iteration error between the adjacent iterations; (c) the optimization solution trend graph.

**C) Classification stability analysis based on LPML-SNMF-based feature learning**

For testing the performance of the proposed integrated ISSRC model when the test data is not balanced in each category, the experiments on Breast-2 dataset are done. In order to verify that the feature learning method LPML-SNMF alleviates the problem of classification unstable, we compared the integrated ISSSRC model with DIF+ISSRC. We fix the total number of the test samples as 20 and change the number of tumors and normals. Fig. 13(a) gives the classification result at different ratios of tumor and normal in the test set, where red curve denotes integrated ISSRC and blue one is DIF+ISSRC, the values on the curves are the ratios of tumor to normal.

In addition to the commonly used classification accuracy, ERR is also adopted into the comparison.

$$ERR = \frac{ER_1 - ER_2}{ER_1} \times 100\%,$$

where $ER_2$ the error rate of highest recognition result on the same method, $ER_1$ is the error rate of other recognition result on the same method, and ERR is denoted by a notion ↓. Fig.13 (b) and Table 10 are ERR values of different ratios of tumor to normal in test samples. Fig. 13 (b) is rose figure of ERR, the smaller the ERR value is, the more concentrated the rose figure is. Experiments are given in the Fig. 13, which shows that: (1) the category-imbalance does affect the classification results, and the classification accuracies of category-balance are superior to category-imbalance. (2) the optimal classification accuracy is achieved when the numbers of samples are balanced. (3) LPML-SNMF-based feature learning makes the classification more stable regardless of whether the category of the test sample is in equilibrium or not.

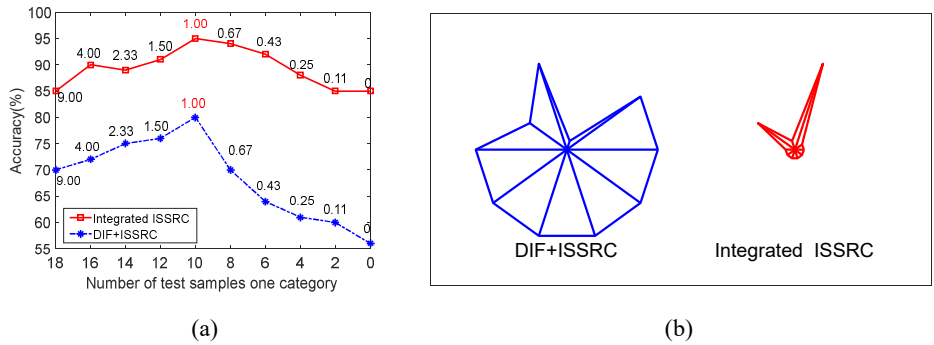

(a)          (b)

Fig. 13. Comparison of classification accuracies when the test data is not balanced in each category. (a) line chart of recognition rate, (b) rose figure of ERR.

Table 10 Classification ERR based on different methods on Breast-2 dataset.

| Methods | Ratio of tumor to normal in the test samples | | | | | | | | | |
| --- | --- | --- | --- | --- | --- | --- | --- | --- | --- | --- |
| | 9.00 | 4.00 | 2.33 | 1.50 | 1.00 | 0.67 | 0.43 | 0.25 | 0.11 | 0 |
| **Integrated ISSRC** | **11.76** | **5.56** | **6.74** | **4.40** | **0** | **1.06** | **3.26** | **7.95** | **11.76** | **11.76** |
| DIF+ISSRC | 14.29 | 11.11 | 6.67 | 5.26 | 0 | 14.29 | 25.00 | 31.15 | 33.33 | 42.86 |

D) **Stability of the integrated ISSRC**

By taking full advantage of the information embedded in test samples, the integrated ISSRC can relieve the problem of insufficient training samples. The performance of SRC and the proposed integrated ISSRC are compared by reducing the number of training samples. In order to verify the stability of the integrated ISSRC, SRC is also based on DIF-based gene selection and LPML-SNMF-based feature learning, without confusion, it is called the integrated SRC. The percentage of the training samples is decreased from 90% to 10%. From Fig. 14, it can be seen that the integrated SRC and the integrated ISSRC reach the similar results when the number of the training samples is more than 80% percentage. With decreasing the number of the training samples, classification accuracy of the integrated SRC will soon lower than the integrated ISSRC. On the whole, the integrated ISSRC performs more stable than the integrated SRC, especially when there are few training samples.

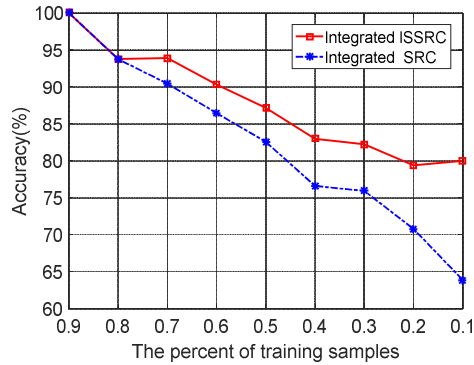

Fig. 14. Comparison of accuracies with decreasing training samples on Breast-2 dataset.

## E) Comparison of the classical classification methods

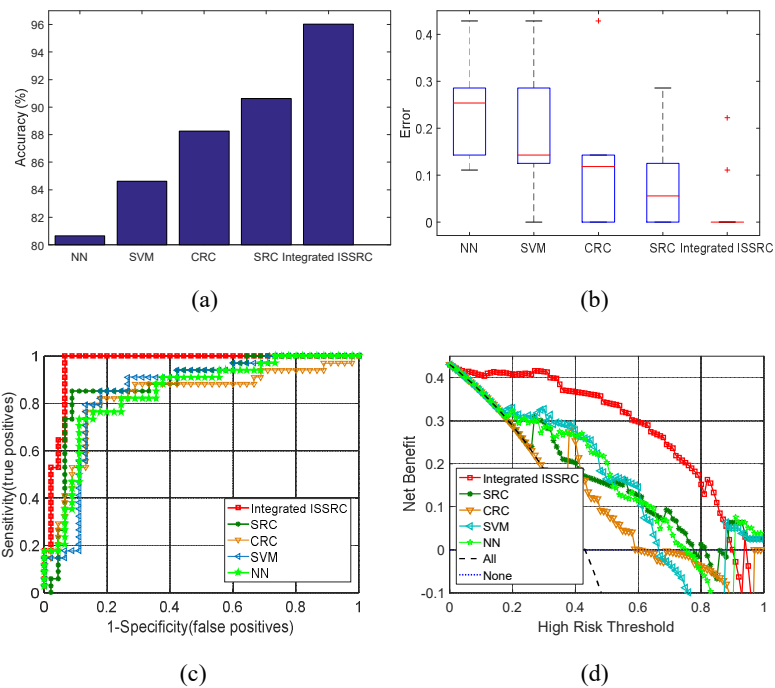

Fig. 15. Comparison of different classification methods on Breast-2 dataset. (a) Histogram for accuracy. (b) box plots for error rates, (c) ROC analysis, (d) DCA analysis.

The performance of the integrated ISSRC is compared with those of NN [52], SVM [6], CRC [13] and SRC [7]. Fig.15 (a) shows the accuracies of different classifiers in the same environment, when comparing the five columns of classification results from Fig.15 (a), it can be seen that the classification accuracy based on the integrated ISSRC is higher than the other methods. In order to give more intuitive comparison of different classifiers, box plots of error rates and ROC analysis are shown in Fig.15. Fig.15 (b) illustrates that NN, SVM,

CRC, SRC and the integrated ISSRC achieve average error rates (red line), and thereby showing that ISSRC has the smallest error rate. Fig.15 (c) gives the ROCs of different classifiers. Specifically speaking, the AUC values, corresponding to ROC, of NN, SVM, CRC, SRC and the integrated ISSRC, are 0.8353, 0.8895, 0.8915, 0.9059 and 0.9458, respectively. DCA analysis is shown in Fig.15 (d), the integrated ISSRC has the highest DCA values, where the higher the DCA is, the smaller the loss of the model is. As can be seen from Fig.14, all the four indexes show that integrated ISSRC is better than other methods.

**F) Comparison of with state-of-the-art methods**

In addition to these classical classifiers, our method has also compared with the latest published classification results on the same breast dataset [48-49, 54-57], and some other state-of-the-art SRC methods, including PFSRC [14], RRC_L1 [43] and RRC_L2 [43].

Table 11 shows that, on the Breast-2 (77) Breast-2(97), classification accuracy of our method higher than those of in the latest published results given in the same dataset and same environment. Especially, on the Breast-2 (97) dataset, classification accuracy of our method achieves 94%, 14%, 8.85% and 6.6% higher than those of in the three latest published results given in [55], [47] and [56] in the same dataset and same environment. And on the Breast-2(77) dataset, the accuracies of our method achieves 94.92% and of increases about 14 percent. The corresponding classification ERR drops about 74 percent.

Table 11 Classification performance with the latest published results on different datasets

| Experiments | Methods | Accuracy (%) |
|---|---|---|
| Breast-2(77) dataset | | |
| Deng et al.(2013) [48] | GRRF-RF | 65.50 |
| Zheng et al.(2017) [57] | CAP-SQDA | 80.00 |
| Fan et al.(2015) [54] | IIS-SQDA | 80.03 |
| **Our paper** | **Integrated ISSRC** | **94.92** |
| Breast-2(97) dataset | | |
| Jiang et al.(2017) [55] | DLPD | 80.00 |
| Younsi et al.(2016) [49] | αRSSE | 85.15 |
| Su et al.(2017) [56] | K-S test-CFS | 87.40 |
| **Our paper** | **Integrated ISSRC** | **94.00** |

Table 12 Classification accuracies of different methods

| Methods | Breast-2 | Breast-2(97) | Breast-2(77) |
|---|---|---|---|
| RRC_L1 [43] | 89.38 | 84.31 | 85.12 |
| RRC_L2 [43] | 88.59 | 85.44 | 86.55 |
| PFSRC [14] | 88.27 | 84.32 | 85.44 |
| **Integrated ISSRC** | **96.03** | **94.00** | **94.92** |

RRC_L1 and RRC_L2 are the recently proposed SRC method, RRC coding model with the L1 and L2 constraints, respectively [43]. PFSRC is another improved SRC method proposed by our team for face recognition [14]. Table 12 shows the results on Breast-2, Breast-2(97) and Breast-2(77) datasets under the same experimental setting. One can observe that our method has a significant advantage, which embodies good recognition results in the prognosis evaluation datasets.

**3.4.4 Analysis of candidate's pathogenic genes**

Tumor characteristics or morphology is very likely to have some relationships with gene expression data. To find out these relationships, the information genes selected based on DIF were further subjected to biological analysis. Our method has achieved good results on postoperative metastasis dataset. The information gene based on DIF selection whether a gene that causes cancer metastasis after treatment? Therefore, the information genes selected based on DIF were further subjected to biological analysis. Identifying candidate's pathogenic genes is important because it can be a biomarker of the candidate's pathogenic genes and it is helpful to auxiliary diagnosis. As shown in subsection 2.1.2, candidate's pathogenic genes can be selected by the proposed DIF index. In this subsection, survival curve analysis is given to a further understanding of its biological meaning. The aim is to study whether these candidate pathogenic genes are used as biomarkers for postoperative metastasis diagnosis. After examining these survival-associated variables, we find that the selected information genes are indeed biologically different foe postoperative metastasis.

Table 13 shows the basic biological attributes of the 10 information genes selected by DIF. Since there are two genes that do not contain the gene name and description, only 8 genes have been analyzed.

Table 13 Some candidate's pathogenic genes and their biological properties for classification on Breast-2 dataset

| Index No. of selected genes | Gene accession number | Gene description |
|---|---|---|
| NM_000286 | PEX12 | peroxisomal biogenesis factor 12 |
| AL080059 | TSPYL5 | TSPY like 5 |
| NM_014968 | PITRM1 | pitrilysinmetallopeptidase 1 |
| AF052087 | CACTIN | cactin, spliceosome C complex subunit |
| NM_003239 | TGFB3 | transforming growth factor beta 3 |
| U45975 | INPP5J | inositol polyphosphate-5-phosphatase J |
| NM_001685 | ATP5J | ATP synthase, H+ transporting, mitochondrial Fo complex subunit F6 |
| NM_019028 | ZDHHC13 | zinc finger DHHC-type containing 13 |

In order to check the quality of DIF-based candidate's pathogenic genes, the expression profiles of these genes for the opposite category are analyzed. For comparison, irrelevant genes chosen randomly are presented. Fig.16 illustrates the two exemplary expression levels of the patients for the candidate's pathogenic genes (TSPYL5 and ATP5J) listed in Table 13 and one irrelevant gene (PTPN1). In Fig.16, the red curve denotes the gene expression levels of the 45 normal samples and the blue curve expresses the gene expression levels of the 34 tumor samples.

The horizontal straight lines indicate the mean values of gene expression levels in the corresponding class. In the case of the pathogenic genes (Figs.16 (a) and (b)), the difference of the mean values is large. For the irrelevant gene (Figs.16 (c)), the difference of the mean values is 0.0132. Moreover, considerable fluctuation can be seen between the binary-category and the irrelevant genes in terms of standard deviation (std). It is implied that gene expression levels of candidate's pathogenic genes are indeed different between normals and patients, while irrelevant genes are normally identical. Therefore, candidate's pathogenic genes can be used to effectively distinguish patients and normals.

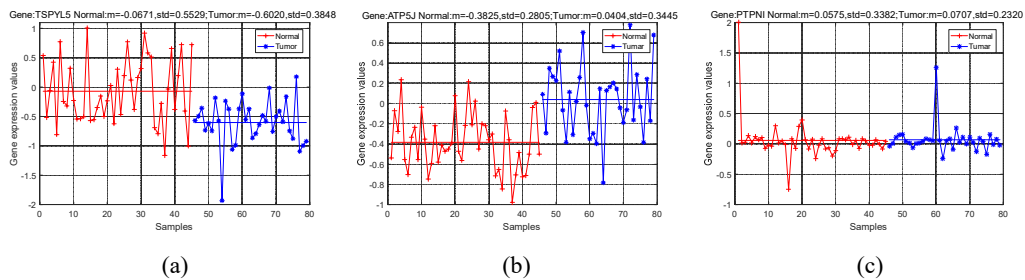

(a)        (b)        (c)

Fig. 16. Comparison of expression levels for candidate's pathogenic genes and irrelevant genes. The red and blue curves correspond to normals and tumors, respectively. (a, b) are pathogenic genes (TSPYL5 and ATP5J), and (c) is the irrelevant genes (PTPN1).

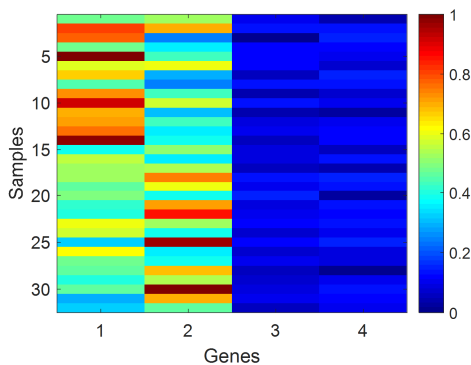

Fig.17. Heatmap of the expression levels for the candidate's pathogenic genes (TSPYL5 and ATP5J, the first two columns) and the irrelevant genes (PTPN1 and ATP2C2-AS1, the last two columns).

Fig.17 shows the heatmap of gene expression levels in two candidate's pathogenic genes (TSPYL5 and ATP5J, the first two columns) and two irrelative genes (PTPN1 and ATP2C2-AS1, the last two columns). It can be seen that the expression levels of the candidate's pathogenic genes have an obvious difference, while there are very similar expression levels in irrelative genes.

Kaplan–Meier estimator is used for patient stratification, and $p$ value is calculated with the log-rank test, where $p < 0.05$ is considered significant. For the 10 candidate's pathogenic genes selected by DIF, we further plot Kaplan-Meier curve by analyzing survival curves and the corresponding Log-Rank $p$ values on website http://www.oncolnc.org and http://ualcan.path.uab.edu/index.html. Fig.18 indicates that TSPYL5 ( $p = 0.0362$ ) is anti-oncogene, PITRM1 ( $p = 0.0066$ ) and ATP5J ( $p = 0.0104$ ) are proto-oncogenes.

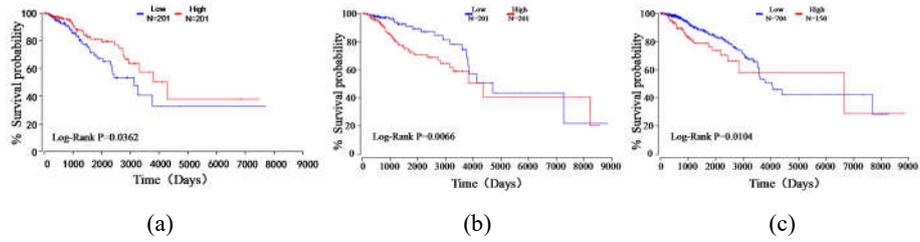

Fig. 18. Kaplan-Meier survival curves of ( $p < 0.05$ ). (a) TSPYL5, (b) PITRM1, (c) ATP5J.

In order to further analyze the selected pathogenic genes, find out whether the specific biological functions of these genes in NCBI and related materials have biological significance for breast tumor. The specific biological information of several genes is given below. Some genes from the final candidate subset for Breast-2 data are shown in Table 13, which are believed to be closely related to Breast tumor. Gene TSPYL5 has been turned out to be associated with Breast tumor in clinical and some pathogenic genes also emerged in the study of others.

Gene TSPYL5 specific biological description: TSPYL5 knockdown decreased, and overexpression increased aromatase (CYP19A1) expression in MCF-7 cells, LCLs, and adipocytes through the skin/adipose (I.4) promoter. TSPYL5 induced CYP19A1 expression and that of many other genes. In summary, genome-wide significant SNPs in TSPYL5 were associated with elevated plasma E2 in postmenopausal breast tumor patients.

Gene ATP5J specific biological description: Mitochondrial ATP synthase catalyzes ATP synthesis, utilizing an electrochemical gradient of protons across the inner membrane during oxidative phosphorylation. Alternatively spliced transcript variants encoding different isoforms have been identified for this gene.

PITRM1: The protein encoded by this gene is an ATP-dependent metalloprotease that degrades post-cleavage mitochondrial transit peptides. Genetic variation in the hPreP gene PITRM1 may potentially contribute to mitochondrial dysfunctions [58].

## 4 Conclusions

In this paper, an integrated ISSR-based tumor classification framework is proposed based on the intrinsic characteristics of microarray gene expression data. The proposed DIF can adaptively select the candidate pathogenic genes, which are consistent with the actual clinical needs and has important biological significance. The LPML-SNMF-based feature learning complements the advantages of deep learning and NMF. The integrated ISSRC is effective and stable, even there are few training samples or the data are unbalanced. Moreover, the integrated ISSRC can effectively identify whether there is a tumor, which kind of tumor, and whether metastasis occurs after surgery.

There remain some interesting questions. One is combing gene network analysis with single gene analysis.

The other is how to further optimize the model, such as adding more targeted prior information as regular terms, considering mixed driven of unlabeled data and model.

## Acknowledgements

The authors would like to thank https://tumorgenome.nih.gov/ for their breast datasets. We also thank Prof. Bingsheng He and Yunhai Xiao for their optimization suggestion, Dr. Yue Li, Pei Wang and Weifeng Yue for bioinformatics' suggestion, Prof. Dexing Kong and Tony W. H. Shen for revision suggestion, Xiaoying Jiang, Chenxi Tian and Lei Sun for their discussion. This work was supported in part by NSF of China (41771375, 11701144), NSF of US (DMS1719932), NSF of Henan Province (162300410061) and Project of Emerging Interdisciplinary (xxjc20170003).

## References

[1] E. T. Liu, C. Sotiriou, Defining the galaxy of gene expression in breast cancer, breast cancer research, 4 (2002) 141-144.

[2] X. B. Zhou, X. D. Wang, E. R. Dougherty, Nonlinear probit gene classification using mutual information and wavelet-based feature selection, Journal of Biological Systems, 12 (2004) 371-386.

[3] L. J. van, H. Dai, M. J. van, Y. D. He, Gene expression profiling predicts clinical outcome of breast cancer, Nature, 415 (2002) 530-536.

[4] L. Breiman, Random forest, Machine Learning, 45 (2001) 5-32.

[5] D. P. Bertsekas, J. N. Tsitsiklis, Neural networks for pattern recognition, Agricultural Engineering International the Cigr Journal of Scientific Research & Development Manuscript Pm, 12 (1995) 1235 – 1242.

[6] T. S. Furey, N. Cristianini, N. Duffy, D. W. Bednarski, M. Schummer, D. Haussler, Support vector machines classification and validation of cancer tissue samples using microarray expression data, Bioinformatics, 16 (2000) 906-914.

[7] J. Wright, A. Ganesh, Z. Zhou, Y. Ma, Robust face recognition via sparse representation, IEEE Transactions on Pattern Analysis and Machine Intelligence, 31 (2009) 210–227.

[8] X. Hang, F. X. Wu, Sparse representation for classification of tumors using gene expression data, Journal of Biomedicine & Biotechnology, 2009 (2014) 403689.

[9] C. H. Zheng, L. Zhang, T. Y. Ng , S. C. Shiu, D. S. Huang, Metasmple-based sparse representation for tumor classification, IEEE/ACM Transactions on Computational Biology & Bioinformatics, 8 (2011) 1273-82.

[10] M. K. Khormuji, M. Bazrafkan, A novel sparse coding algorithm for classification of tumors based on gene expression data, Medical & Biological Engineering & Computing, 54 (2016) 869.

[11] B. Gan, C. H. Zheng, J. Zhang, H. Q. Wang, Sparse representation for tumor classification based on feature extraction using

latent low-rank representation, Biomed Research International, 10 (2014) 63-68.

[12] C. H. Zheng, L. Zhang, T. Y. Ng, S.C.K. Shiu, D.S. Huang, Metasample-Based Sparse Representation for Tumor Classification, IEEE Trans. TCBB, 2011 , (2011) 1273.

[13] L. Zhang, M. Yang, X. Feng, Sparse representation or collaborative representation: which helps face recognition?, IEEE International Conference on Computer Vision, 2011 (2012) 471-478.

[14] X. H. Yang, F. Liu, L.Tian, H. F. Li, X. Y. Jiang, Pseudo-full-space representation based classification for robust face recognition, Signal Processing: Image Communications, 60 (2018) 64-78.

[15] L. J. van't Veer, H. Dai, M. J. V. D. Vijver, Y. D. He, Expression profiling predicts poor outcome of disease in young breast cancer patients, European Journal of Cancer, 37 (2001) S271–S271.

[16] H. Q. Wang, H. S. Wong, D. S Huang, J. Shu, Extracting gene regulation information for cancer classification, Pattern Recognition, 40(2007) 3379-3392.

[17] S. Dudoit, J. Fridlyand, T. P. Speed, Comparison of discrimination methods for the classification of tumors using gene expression data, Journal of the American Statistical Association, 97 (2002) 77-87.

[18] T. R. Golub, D. K. Slonim, P. Tamayo, C. Huard, M. Gaasenbeek, J. P. Mesirov, Molecular classification of cancer: class discovery and class prediction by gene expression monitoring, Science, 286 (1999) 531-537.

[19] Z. Y. Algamal, H. T. M. Ali, An efficient gene selection method for high-dimensional microarray data based on sparse logistic regression, Electronic Journal of Applied Statistical Analysis, 10 (2017) 242-256.

[20] R. Ruiz, J. C. Riquelme, J. S. Aguilar-Ruiz, Incremental wrapper-based gene selection from microarray data for cancer classification, Pattern Recognition, 39(2006) 2383-2392.

[21] J. Xie, M. Wang, H. U. Qiufeng, The differentially expressed gene selection algorithms for unbalanced gene datasets by maximize the area under ROC, Journal of Shaanxi Normal University, 2017.

[22] A. P. Bradley, The use of the area under the ROC curve in the evaluation of machine learning algorithms, Pattern Recognition, 30 (1997) 1145-1159.

[23] A. J. Vickers, E. B. Elkin, Decision curve analysis: a novel method for evaluating prediction models, NIH-PA Author Manuscript, 26 (2006) 565–574.

[24] D. D. Lee, H. S. Seung, Learning the parts of objects by non-negative matrix factorization, Nature, 401 (1999) 788-791.

[25] N. Gillis, F. Glineur, Using under approximations for sparse nonnegative matrix factorization, Pattern Recognition, 43 (2010) 1676–1687.

[26] S. Wild, J. Curry, A. Dougherty, Improving non-negative matrix factorizations through structured initialization, Pattern Recognition, 37(2004) 2217-2232.

[27] C. Boutsidis, E. Gallopoulos, SVD based initialization: A head start for nonnegative matrix factorization, Pattern Recognition,

41(2008) 1350-1362.

[28] N. Yuvaraj, P. Vivekanandan, An efficient SVM based tumor classification with symmetry non-negative matrix factorization using gene expression data, International Conference on Information Communication and Embedded Systems (ICICES), (2013) 761-768.

[29] C. H. Zheng, T. Y. Ng, L. Zhang, C. K. Shiu, H. Q. Wang, Tumor classification based on non-negative matrix factorization using gene expression data, IEEE Transactions on Nanobioscience, 10 (2011) 86-93.

[30] D. Tu, L. Chen, G. C. Chen, Y. Wu, J. C. Wang, Hierarchical online NMF for detecting and tracking topic hierarchies in a text stream, Pattern Recognition, 76(2018) 203-214.

[31] A. Cichocki, R. Zdunek, Multilayer nonnegative matrix factorization, Electronics Letters, 42 (2006) 947-948.

[32] P. O. Hoyer, Non-negative sparse coding, IEEE Workshop on Neural Networks for Signal Processing, (2002) 557-565.

[33] Y. Lecun, Y. Bengio, G. Hinton, Deep learning, Nature, 521(2015) 436.

[34] P. Guillen, J. Ebalunode, Cancer classification based on microarray gene expression data using deep learning, International Conference on Computational Science and Computational Intelligence (CSCI), 2016.

[35] Z. Y. Han, B. Z. Wei, Y. J. Zheng, Y. L. Yin, K. J. Li, S. Li, Breast cancer multi-classification from histopathological images with structured deep learning, Science Reports, 7 (2017).

[36] Z. B. Xu, J. Sun, Model-driven deep-learning, National Science Review, 5 (2018) 22-24.

[37] B. Efron, T. Hastie, I. Johnstone, R.Tibshirani, Least angle regression, Annals of Statistics, 32 (2004) 407-451.

[38] M. R. Hestenes, Multiplier and gradient methods, Journal of Optimization Theory &Applications, 4 (5) (1969) 303-320.

[39] Y. H. Xiao, L. Chen, D. Li, A generalized alternating direction method of multipliers with semi-proximal terms for convex composite conic programming, Mathematical Programming Computation, 1 (2018) 1-23.

[40] U. Alon, N. Barkai, D.A. Notterman, K. Gish, S. Ybarra, D. Mack, and A.J. Levine," Broad patterns of gene expression revealed by clustering analysis of tumor and normal colon tissues probed by oligonucleotide arrays," Proc. Nati. Acad. Sci.., vol. 96, pp.6745-6750, 1999.

[41] P. Tamayo, "Diffuse large B-cell lymphoma outcome prediction by gene expression profiling and supervised machine learning," Nat. Med., Vol. 8, No. 1. pp.68-74, 2002.

[42] S.A. Armstrong, J.E. Staunton, L.B.Silverman, R. Pieters, M.L. den Boer, M.D. Minden, S.E. Sallan, E.S. Lander, T.R. Golub, and S.J. Korsmeyer, "MLL translocations specify a distinct gene expression profile that distinguishes a unique leukemia," Nat. Gene., vol. 30, pp. 41-47, 2002.

[43] M. Yang, L. Zhang, J. Yang, D. Zhang, Regularized robust coding for face recognition, IEEE Transactions, Image Processing, 22 (2012) 1753-1766.

[44] J. X. Liu, Y. Xu, C. H. Zheng, H. Kong, Z. H. Lai, RPCA-based tumor classification using gene expression data, IEEE Trans.

TCBB, 12 (2015) 964-970.

[45] V. García, J. S. Sánchez, Mapping microarray gene expression data into dissimilarity spaces for tumor classification, Information Sciences, 294(2015) 362-375.

[46] M. Dettling, P. Bühlmann, BagBoosting for tumor classification with gene expression data, Bioinformatics, 20 (2004) 1061-1069.

[47] B. Gan, C. H. Zheng, J. X. Liu, Metasample-based robust sparse representation for tumor classification, Engineering, 05 (2016) 78-83.

[48] H. Deng, G. Runger, Gene selection with guided regularized random forest, Pattern Recognition, 46(2013) 3483-3489.

[49] R. Younsi, A. Bagnall, Ensembles of random sphere cover classifiers, Pattern Recognition, 49(2016) 213-225.

[50] Y. Piao, M. Piao, K. Park, K. H. Ryu, An ensemble correlation-based gene selection algorithm for cancer classification with gene expression data, Bioinformatics, 28 (2012) 3306-3315.

[51] J. H. Hong, S. B. Cho, Gene boosting for cancer classification based on gene expression profiles, Pattern Recognition, 42(2009) 1761-1767.

[52] T. W. Cover, P. E. Hart, Nearest neighbor pattern classification, IEEE Transactions On Information Theory, 13 (1967) 21-27.

[53] W. Deng, J. Hu, and J. Guo, Extended SRC: undersampled face recognition via intraclass variant dictionary, IEEE Trans. Pattern Anal, 34(2012) 1864-1870.

[54] Y. Fan, Y. Kong, D. Li, Z. Zheng, Innovated interaction screening for high-dimensional nonlinear classification, Annals of Statistics, 43 (2015) 1243-1272.

[55] B. Jiang, Z. Chen, C. Leng, Dynamic linear discriminant analysis in high dimensional space, arXiv: 1708.00205, 2017.

[56] Q. Su, Y. Jiang, F. Chen, W. C. Lu, A cancer gene selection algorithm based on the K-S test and CFS, Biomed Research International, 2017.

[57] D. Zheng, J. Jia, X. Fang, X. Guo, Main and interaction effects selection for quadratic discriminant analysis via penalized linear regression, arXiv: 1702.0457, 2017.

[58] C. M. Pinho, B. F. Bjork, N. Alikhani, Genetic and biochemical studies of SNPs of the mitochondrial a beta-degrading protease, Neuroscience Letters, 469 (2010) 204-208.

[59] G.H. Golub, C.F.V. Loan, Matrix computations, Johns Hopkins University, Press, Balt imore, MD, USA, (1996) 242-243.

# Appendix A

Firstly, $H_1$ is optimized for a given basis $W_1$. Since the objective function (9a) is quadratic with respect to $H_1$, and the feasible region is of convex type, it is guaranteed that there exists a local minimum. To address this problem, [32] has given an iterative update rule.

$$H_1^{k+1} = H_1^k .* (W_1^k)^T V ./ ((W_1^k)^T W_1^k H_1^k + \lambda_1). \tag{20}$$

In order to satisfy the nonnegative constraints $H_1 > 0$, projection operators $\hat{h}_{1ij}$ can be constructed as follows, and they are applied on the optimization solutions.

$$\hat{h}_{1ij} = \begin{cases} h_{1ij}, & if\ h_{1ij} \geq 0, \\ 0, & otherwise. \end{cases} h_1 \in H_1^{k+1}, \tag{21}$$

where $h_1$ is the vectors of the matrices $H_1^{k+1}$, $h_{1ij}$ is the elements in $h_1$. Eq. (21) denotes that all elements in matrix $H_1^{k+1}$ is projected into a non-zero space.

Then $H_1$ is fixed and the basis $W_1$ is optimized.

$$W_1^{k+1} = W_1^k - \mu_1 (W_1^k H_1^{k+1} - V)(H_1^{k+1})^T, \tag{22}$$

where $\mu_1 > 0$ is an iteration step. In order to satisfy the nonnegative constraints $W_1 > 0$, projection operators $\hat{w}_{1ij}$ can be constructed as follows, and they are applied on the optimization solutions.

$$\hat{w}_{1ij} = \begin{cases} w_{1ij}, & if\ w_{1ij} \geq 0, \\ 0, & otherwise. \end{cases} w_1 \in W_1^{k+1}, \tag{23}$$

where $w_1$ is the vectors of the matrices $W_1^{k+1}$, $w_{1ij}$ is the elements in $w_1$. Eq. (23) denotes that all elements in matrix $W_1^{k+1}$ is projected into a non-zero space.

Similarly, the model in Eq. (9b) can be also optimized and $H_2$ and $W_2$ are updated by alternating iterations

$$H_2^{k+1} = H_2^k .* (W_2^k)^T H_1 ./ ((W_2^k)^T W_2^k H_2^k + \lambda_2), \tag{24}$$

$$\hat{h}_{2ij} = \begin{cases} h_{2ij}, & if\ h_{2ij} \geq 0, \\ 0, & otherwise. \end{cases} h_2 \in H_2^{k+1}, \tag{25}$$

$$W_2^{k+1} = W_2^k - \mu_2 (W_2^k H_2^{k+1} - H_1)(H_2^{k+1})^T, \tag{26}$$

$$\hat{w}_{2ij} = \begin{cases} w_{2ij}, & \text{if } w_{2ij} \geq 0, \\ 0, & \text{otherwise.} \end{cases} \quad w_2 \in W_2^{k+1}. \tag{27}$$

## Appendix B

***Proof***: In order to discuss the value of $\dfrac{\|\alpha_j - \alpha_i\|_2}{\|\alpha_i\|_2}$, we need to find the relationship between $\alpha_i$ and $\alpha_j$. Let $\alpha_i(t)$ is continuously differentiable for all $t \in [0, \varepsilon]$, where $\alpha_i = \alpha_i(0)$ and $\alpha_j = \alpha_i(\varepsilon)$. Let $\alpha_i(t)$ do the Taylor expansion at $t = 0$: $\alpha_{i(t)} = \alpha_i(0) + \varepsilon \alpha_i'(0) + O(t^2)$. We have $\alpha_j = \alpha_i + \varepsilon \alpha_i'(0) + O(\varepsilon^2)$ when $t = \varepsilon$. Then

$$\frac{\|\alpha_j - \alpha_i\|_2}{\|\alpha_i\|_2} = \varepsilon \frac{\|\alpha_i'(0)\|_2}{\|\alpha_i\|_2} + O(\varepsilon^2). \tag{28}$$

In order to obtain $\|\alpha_i'(0)\|_2$, similar to Theorem 5.3.1 in [59], one can construct $\left(H_2^{test} + tf\right)^{\mathrm{T}}\left(H_2^{test} + tf\right)\alpha_i(t)$, where $f = \Delta\left(H_2^{test}\right) / \varepsilon$, then

$$\left(H_2^{test} + tf\right)^{\mathrm{T}}\left(H_2^{test} + tf\right)\alpha_i(t) = \left(H_2^{test} + tf\right)^{\mathrm{T}}\left((h_2^{train})_i + t\Delta\left(H_2^{test}\right)\alpha_i(t)/\varepsilon\right).$$

Let $E = \Delta\left((h_2^{train})_i\right)/\varepsilon$, then

$$\left(H_2^{test} + tf\right)^{\mathrm{T}}\left(H_2^{test} + tf\right)\alpha_i(t) = \left(H_2^{test} + tf\right)^{\mathrm{T}}\left((h_2^{train})_i + tE\right). \tag{29}$$

In order to bound $\|\alpha_i'(0)\|_2$, one can take the derivative of Eq. (29) and set $(h_2^{train})_j$,

$f^T H_2^{test} \alpha_i + (H_2^{test})^T f \alpha_i + H_2^{test}(H_2^{test})^T \alpha_i'(0) = (H_2^{test})^T E + f^T (h_2^{train})_i$ i.e.,

$$\alpha_i'(0) = \left((H_2^{test})^{\mathrm{T}} H_2^{test}\right)^{-1} (H_2^{test})^{\mathrm{T}}\left(E - f\alpha_i\right) + \left((H_2^{test})^{\mathrm{T}} H_2^{test}\right)^{-1} f^{\mathrm{T}}\left((h_2^{train})_i - H_2^{test}\alpha_i\right). \tag{30}$$

By singular value decomposition theorem [59], we have $rank\left(H_2^{test} + tf\right) = k$ for all $t \in [0, \varepsilon]$, where $\left\|\Delta\left(H_2^{test}\right)\right\|_2 \leq \varphi_k\left(H_2^{test}\right)$ ($\varphi_k\left(H_2^{test}\right)$ is the largest singular value of $H_2^{test}$). Then

$$\|f\|_2 = \|\Delta\left(H_2^{test}\right)/\varepsilon\|_2 \leq \varphi_k\left(H_2^{test}\right) \leq \|H_2^{test}\|_2,$$

and $\|E\|_2 = \|\Delta\left((h_2^{train})_i\right)/\varepsilon\|_2 \leq \|(h_2^{train})_i\|_2$.

By substituting Eq. (30) result into Eq. (28), taking norms, the inequality can be obtained,

$$\frac{\|\alpha_j - \alpha_i\|_2}{\|\alpha_i\|_2} \leq \varepsilon \left\{ \|H_2^{test}\|_2 \cdot \|((H_2^{test})^T H_2^{test})^{-1}(H_2^{test})^T\|_2 \cdot \left( \frac{\|(h_2^{train})_i\|_2}{\|H_2^{test}\|_2 \|\alpha_i\|_2} + 1 \right) \right.$$
$$\left. + \frac{\rho_{LS}}{\|H_2^{test}\|_2 \|\alpha_i\|_2} \cdot \|H_2^{test}\|_2^2 \cdot \|((H_2^{test})^T H_2^{test})^{-1}\|_2 \right\} + O(\varepsilon^2).$$

Since $(H_2^{test})^T(H_2^{test}\alpha_i - (h_2^{train})_i) = 0$, $H_2^{test}\alpha_i$ is orthogonal to $H_2^{test}\alpha_i - (h_2^{train})_i$, it is also known that $\|(h_2^{train})_i - H_2^{test}\alpha_i\|_2^2 + \|H_2^{test}\alpha_i\|_2^2 = \|(h_2^{train})_i\|_2^2$, then $\|H_2^{test}\|_2^2 \cdot \|\alpha_i\|_2^2 \geq \|(h_2^{train})_i\|_2^2 - \rho_{LS}^2$.

The relationship between $\alpha_i$ and $\alpha_j$ will be

$$\frac{\|\alpha_j - \alpha_i\|_2}{\|\alpha_i\|_2} \leq \varepsilon \left\{ \kappa_2(H_2^{test}) \left( \frac{1}{\cos(\theta)} + 1 \right) + \kappa_2(H_2^{test})^2 \frac{\sin(\theta)}{\cos(\theta)} \right\} + O(\varepsilon^2).$$

**Appendix C**

The integrated ISSR model in Eq. (11) can be rewritten as Eq. (13) (in subsection 2.2)

$$\min_{\alpha,b} \|h_2^{train} - H_2^{test}\alpha\|_2^2 + \lambda \|b\|_1 \quad s.t. \quad \alpha - b = 0. \tag{13}$$

For $h_2^{train} \in R^{r_2 \times 1}$, $H_2^{test} \in R^{r_2 \times k}$, the augmented Lagrangian function of (13) is defined as,

$$L_\sigma(\alpha,b;\eta) = \|h_2^{train} - H_2^{test}\alpha\|_2^2 + \lambda \|b\|_1 + \langle \eta, \alpha - b \rangle + \frac{\sigma}{2}\|\alpha - b\|_2^2, \tag{14}$$

Let $\sigma > 0$ be the penalty parameter, and $\eta \in R^{k \times 1}$ be the Lagrange multiplier, $\langle \cdot, \cdot \rangle$ denotes the inner product.

The GsADMM scheme takes the following form

$$\begin{cases} \alpha^k = \arg\min_\alpha L_\sigma(\alpha, \tilde{b}^k; \tilde{\eta}^k) + \frac{1}{2}\|\alpha - \tilde{\alpha}^k\|_K^2, & (a) \\ \eta^k = \tilde{\eta}^k + \sigma(\alpha^k - \tilde{b}^k), & (b) \\ b^k = \arg\min_b L_\sigma(\alpha^k, b; \eta^k) + \frac{1}{2}\|b - \tilde{b}^k\|_T^2, & (c) \\ \tilde{w}^{k+1} = \tilde{w}^k + \rho(w^k - \tilde{w}^k), & (d) \end{cases} \tag{15}$$

where $w^k = (\alpha^k, b^k, \eta^k)$, $\alpha^{-1} = \tilde{\alpha}^0$, $b^{-1} = \tilde{b}^0$, $K: R^{k \times 1} \to R^{k \times 1}$ and $T: R^{k \times 1} \to R^{k \times 1}$ are two semi-proximal matrixes. A more natural choice of the semi-proximal terms is to add $\frac{1}{2}\|\alpha - \tilde{\alpha}^k\|_K$ and $\frac{1}{2}\|b - \tilde{b}^k\|_T$ to the sub-problems for computing the values $\alpha^k$ and $b^k$. For the sake of generality and numerical convenience, the latter variant with only semi-proximal terms is considered. The most adopted values of the variables are used in the proximal terms.

Sub-problem $\alpha^k$ can be approximated by

$$\begin{aligned}
\alpha^k &= \arg\min_{\alpha} \left\| h_2^{train} - H_2^{test}\alpha \right\|_2^2 + \frac{\sigma}{2}\left\| \alpha - \tilde{b}^k + \frac{\tilde{\eta}^k}{\sigma} \right\|_2^2 + \frac{1}{2}\left\| \alpha - \tilde{\alpha}^k \right\|_K^2, \\
&= \arg\min_{\alpha} \left\langle Z^k, \alpha - \tilde{\alpha}^k \right\rangle + \frac{\theta_K}{2}\left\| \alpha - \tilde{\alpha}^k \right\|_2^2 \\
&= \arg\min_{\alpha} \frac{\theta_K}{2}\left\| \alpha - \tilde{\alpha}^k + \frac{Z^k}{\theta_K} \right\|_2^2 = \tilde{\alpha}^k - \frac{Z^k}{\theta_K}
\end{aligned} \quad (31)$$

where $Z^k = (H_2^{test})^T H_2^{test}\tilde{\alpha}^k - (H_2^{test})^T h_2^{train} - \sigma\tilde{b}^k + \sigma\tilde{\alpha}^k + \tilde{\eta}^k$, and $Y^T Y + \sigma I = K$, $\theta_K > \|K\|_F^2$.

Sub-problem $b^k$ can be approximated by

$$\begin{aligned}
b^k &= \arg\min_{b} \lambda\|b\|_1 + \langle \eta^k, \alpha^k - b \rangle + \frac{\sigma}{2}\|\alpha^k - b\|_2^2 + \frac{1}{2}\|b - \tilde{b}^k\|_T^2, \\
&= \arg\min_{b} \lambda\|b\|_1 + \frac{\sigma}{2}\left\|\alpha^k - b + \frac{\eta^k}{\sigma}\right\|_2^2 + \frac{1}{2}\|b - \tilde{b}^k\|_T^2,
\end{aligned}$$

where $T = 0$, then

$$b^k = \arg\min_{b} \lambda\|b\|_1 + \frac{\sigma}{2}\left\|\alpha^k - b + \frac{\eta^k}{\sigma}\right\|_2^2 = S_{\lambda/\sigma}\left(\alpha^k + \frac{\eta^k}{\sigma}\right), \quad (32)$$

where $S$ is a soft threshold function given below

$$S_{\varepsilon}[x] = \begin{cases} x - \varepsilon, & \text{if } x > \varepsilon \\ x + \varepsilon, & \text{if } x < -\varepsilon \\ 0, & \text{otherwise} \end{cases}.$$

---

**Algorithm 2: Optimization of integrated ISSR based on GsADMM**

**Input:** Training samples feature matrix $H_2^{train} = [(h_2)_1^{train}, \cdots, (h_2)_{s_c}^{train}]$, test samples feature matrix $H_2^{test} = [(h_2^{test})_1, \cdots, (h_2^{test})_k]$. Set $\rho \in (0, 2)$ and $\sigma > 0$.

**Initialize:** Initialize $(\tilde{\alpha}^0, \tilde{b}^0, \tilde{\eta}^0) = (0, 0, 0)$, $k = 0$, $Y^T Y + \sigma I = K$, $\theta_K > \|K\|_F^2$.

**Iterate the following processes until convergence**
1) $\alpha^k = \tilde{\alpha}^k - ((H_2^{test})^T H_2^{test}\tilde{\alpha}^k - (H_2^{test})^T h_2^{train} - \sigma\tilde{b}^k + \sigma\tilde{\alpha}^k + \tilde{\eta}^k)/\theta_K$,
2) $\eta^k = \tilde{\eta}^k + \sigma(\alpha^k - \tilde{b}^k)$,
3) $b^k = S_{\lambda/\sigma}(\alpha^k + \eta^k/\sigma)$,
4) $\tilde{w}^{k+1} = \tilde{w}^k + \rho(w^k - \tilde{w}^k)$,
5) $k = k + 1$.

**End while**

**Output** An optimal solution can be obtained.

---

## Appendix D

For a further discussion of the analyzed results, the notations $\alpha_{<i} := (\alpha_1, \cdots, \alpha_{i-1})$, $\alpha_{>i} := (\alpha_{i+1}, \cdots, \alpha_p)$ and similar notations for $b$ are employed. Let $f(\alpha) = \sum_{j=1}^{q} f_j(\alpha_j)$, which is abbreviated as $\Sigma_f$, $g(b) = \sum_{j=1}^{q} g_j(b_j)$, $1 \leq j \leq q$ for $\Sigma_g$.

In [39], two conditions for $\Sigma_f + K$ and $\Sigma_g + T - I$ are needed. So, the following two basic equalities are given first.

$$2\langle u_1 - u_2, G(v_1 - v_2)\rangle = \|u_1 - v_2\|_G^2 - \|u_1 - v_1\|_G^2 + \|u_2 - v_1\|_G^2 - \|u_2 - v_2\|_G^2, \tag{33}$$

$$\begin{aligned}2\langle u_1 - Gu_2\rangle &= \|u_1\|_G^2 + \|u_2\|_G^2 - \|u_1 - u_2\|_G^2 \\ &= \|u_1 + u_2\|_G^2 - \|u_1\|_G^2 - \|u_2\|_G^2,\end{aligned} \tag{34}$$

where $u_1$, $u_2$, $v_1$, $v_2$ are vectors, and $G$ is an arbitrary self-adjoin positive semi-definite linear operator from a space to itself.

Next, let $(\bar{\eta}, \bar{\alpha}, \bar{b})$ be an arbitrary solution to the KKT system (16). For any $(\eta, \alpha, b)$, we denote $\eta_e = \eta - \bar{\eta}$, $\alpha_e = \alpha - \bar{\alpha}$ and $b_e = b - \bar{b}$.

**Proof**: Note that $\rho \in (0,2)$. It is clear to see from Eq. (19) (in subsection 2.3) that $\{\phi_k\}_k \geq 0$ is a nonnegative and monotonically non-increasing sequence. Hence, $\{\phi_k\}$ is also bounded. As a result, the following sequences are bounded,

$$\{\|\eta_e^k + \sigma(1-\rho)\alpha_e^k\|\}, \{\|\tilde{\alpha}^{k+1}\|_K\}, \{\|\tilde{b}^k\|_T\}, \{\|\alpha^k - \tilde{\alpha}^k\|_K\}, \text{ and } \{\|\alpha_e^k\|\}. \tag{35}$$

Moreover, it is known that the following inequalities hold with $k \to \infty$,

$$\|\alpha_e^{k+1}\|_{\Sigma_f} \to 0, \ \|b_e^k\|_{\Sigma_g} \to 0, \ \|\alpha_e^{k+1} - b_e^k\| \to 0, \ \|\tilde{\alpha}^{k+1} - \alpha^{k+1}\|_K \to 0,$$

$$\|\tilde{b}^k - b^k\|_T \to 0, \ \|\alpha^{k+1} - \alpha^{k+1}\|_{\Sigma_f} \to 0 \text{ and } \|\alpha_e^k - \alpha_e^{k+1}\| \to 0. \tag{36}$$

Thus, it can be seen that $\{\|\alpha^k\|_K\}$ is bounded by the fact of $\|\alpha^k\|_K \leq \|\alpha^k - \tilde{\alpha}^k\|_K + \|\tilde{\alpha}^k\|_K$. Consequently, the sequence $\{\|\alpha^k\|_{\Sigma_f + K}\}$ is bounded. Since $\Sigma_f + K > 0$, the sequence $\{\|\alpha^k\|\}$ is also bounded. Similarly, the sequences $\{\|b^k\|_T\}$, $\{\|-b^k\|\}$ and $\{\|b^k\|_{\Sigma_g + T - I}\}$ are both bounded, which leads to the bounded sequence as well. It is implied that the sequence $\{\|b^k\|\}$ is bounded from $\Sigma_g + T - I > 0$. The boundedness of $\{\|\eta_e^k + \sigma(1-\rho)\alpha_e^k\|\}$ and $\{\|\alpha^k\|\}$ further indicate that the sequence $\{\|\eta^k\|\}$ is bounded. The above arguments

have shown that $\{(\eta^k, \alpha^k, b^k)\}$ is a bounded sequence.

Consequently, the sequence $\{(\eta^k, \alpha^k, b^k)\}$ admits at least one convergent subsequence. Assume that $\{(\eta^{k_i}, \alpha^{k_i}, b^{k_i})\}$ is a subsequence of $\{(\eta^k, \alpha^k, b^k)\}$ converging to $(\eta^\infty, \alpha^\infty, b^\infty)$. It follows from Eq. (15b) that

$$\eta^k - K(\alpha^k - \tilde{\alpha}^k) \in \partial f(\alpha^k), \text{ and } \eta^{k+1} - K(\alpha^{k+1} - \tilde{\alpha}^{k+1}) \in \partial f(\alpha^{k+1}), \tag{37}$$

Now, from Eq. (15c) we can get

$$-\eta^{k+1} + (-\eta^k + \eta^{k+1}) + \sigma(\alpha^k - b^k) - T(b^k - \tilde{b}^k) \in \partial g(b^k). \tag{38}$$

It follows from Eqs. (37) and (38) that

$$\begin{cases} \eta^{k_i} - K(\alpha^{k_i} - \tilde{\alpha}^{k_i}) \in \partial f(\alpha^{k_i}), \\ -\eta^{k_i} + \sigma(\alpha^{k_i} - b^{k_i}) - T(b^{k_i} - \tilde{b}^{k_i}) \in \partial g(b^{k_i}). \end{cases} \tag{39}$$

$\lim_{k \to \infty}(\alpha^k - b^k) = 0$ can be obtained from Eq. (36). Taking limit on Eq. (36) and using Eq. (39), one obtains

$$\eta^\infty \in \partial f(\alpha^\infty), \quad -\eta^\infty \in \partial g(b^\infty) \text{ and } \alpha^\infty - b^\infty = 0,$$

which indicates that $(\eta^\infty, \alpha^\infty, b^\infty)$ is a solution to the KKT system (16).

Next we will show that $(\eta^\infty, \alpha^\infty, b^\infty)$ is the unique limit point of the sequence $\{(\eta^k, \alpha^k, b^k)\}$. Without loss of generality, let $(\bar{\eta}, \bar{\alpha}, \bar{b}) = (\eta^\infty, \alpha^\infty, b^\infty)$. Consequently, the sequence $\{\phi_k\}$ itself converges to zero and the $\lim_{k \to \infty} \eta^k = \bar{\eta}$ by the definition of $\phi_k$. Moreover, from $\|\tilde{\alpha}^{k+1} - \alpha^{k+1}\|_K \to 0$ in Eq. (36), it is easy to get $\|\alpha_e^k\|_K \to 0$ as $k \to \infty$. Noting that $\|\alpha_e^k\|_K \to 0$ in Eq. (35) and $\|\alpha_e^k\|_{\Sigma_f} \to 0$ in Eq. (36), we have $\{\|\alpha_e^k\|_{\Sigma_f} + \|\alpha_e^k\|_K + \|\alpha_e^k\|\} \to 0$ as $k \to \infty$. Hence, $\lim_{k \to \infty} \alpha^k = \bar{\alpha}$ under the condition of $\Sigma_f + K > 0$. Finally, from the facts of $\|\alpha_e^{k+1}\| \to 0$, $\|\alpha_e^{k+1} - b_e^k\| \to 0$ in (36), and

$$\|-b_e^k\| \leq \|\alpha_e^{k+1} - b_e^k\| + \|\alpha_e^{k+1}\|,$$

$\|-b_e^k\| \to 0$ can be derived. Since $\|b_e^k\|_T \to 0$ and $\|b_e^k\|_{\Sigma_g} \to 0$ owing to Eq. (36), one can get $\{\|b_e^k\|_{\Sigma_g} + \|b_e^k\|_T + \|-b_e^k\|\} \to 0$ as $k \to \infty$. Therefore, by the fact that $\Sigma_g + T - I > 0$, it is known that $\lim_{k \to \infty} b^k = \bar{b}$.